\documentclass[preprint,12pt]{elsarticle}

\usepackage{amssymb}
\usepackage{amsmath}
\usepackage{graphicx}
\usepackage{algorithm}
\usepackage{algpseudocode}

\usepackage[font=small,labelfont=bf]{caption}
\usepackage{caption} % For captions
\usepackage{subcaption} % For subfigure
\usepackage{float} % For H option
\usepackage{booktabs}
\usepackage{makecell} % for \makecell

\usepackage{rotating} % for \rotatebox
\usepackage{algpseudocode}

\usepackage{makecell}
\usepackage{adjustbox}
\usepackage{multirow}
\usepackage{tabularray}

\DeclareRobustCommand{\method}[1]{{\fontsize{8}{12}\selectfont \textbf{#1}}}

\newtheorem{theorem}{Theorem}

\usepackage{threeparttable}

\usepackage{enumitem}

\usepackage{booktabs}     % For \toprule, \midrule, \bottomrule
\usepackage{tabularx}     % For tabularx environment and column width specification
\usepackage{caption}      % For customizing table captions (optional but recommended)
\usepackage{array}        % For better column formatting (optional)
\usepackage{multirow}     % If using \multirow (currently not used but helpful if you expand)
\usepackage{caption}   % For better control over captions
\usepackage{float}     % For [H] placement option (if needed)

\usepackage{makecell}

\usepackage{longtable}
\usepackage{booktabs} % for \toprule, \midrule, \bottomrule

\usepackage{longtable}
\usepackage{xspace}
\usepackage{booktabs}
\usepackage{pifont} % for checkmark and xmark
\newcommand{\cmark}{\text{\ding{51}}}%
\newcommand{\xmark}{\text{\ding{55}}}%

\newcommand{\modelname}{\textit{OLC-WA}\xspace}

\usepackage{ltablex} % loads longtable + tabularx combo
\keepXColumns
\usepackage{makecell,booktabs,graphicx,array}
\newcolumntype{Y}{>{\raggedright\arraybackslash}X}
\newcommand{\rot}[1]{\rotatebox[origin=c]{90}{#1}}

\usepackage[hyphens]{url}    % allow line breaks at hyphens in URLs
\usepackage[hidelinks]{hyperref} % clickable links, no colors or boxes
\usepackage{xcolor}

\usepackage{rotating} % in preamble

\journal{Expert Systems with Applications}

\begin{document}

\begin{frontmatter}

%% Title, authors and addresses

%% use the tnoteref command within \title for footnotes;
%% use the tnotetext command for theassociated footnote;
%% use the fnref command within \author or \affiliation for footnotes;
%% use the fntext command for theassociated footnote;
%% use the corref command within \author for corresponding author footnotes;
%% use the cortext command for theassociated footnote;
%% use the ead command for the email address,
%% and the form \ead[url] for the home page:
%% \title{Title\tnoteref{label1}}
%% \tnotetext[label1]{}
%% \author{Name\corref{cor1}\fnref{label2}}
%% \ead{email address}
%% \ead[url]{home page}
%% \fntext[label2]{}
%% \cortext[cor1]{}
%% \affiliation{organization={},
%%             addressline={},
%%             city={},
%%             postcode={},
%%             state={},
%%             country={}}
%% \fntext[label3]{}

\title{\modelname: Drift Aware Tuning-Free Online Classification with Weighted Average}
% \title{\modelname Drift Aware Tuning-Free Online Classification with Weighted Average}

% %% use optional labels to link authors explicitly to addresses:
% %% \author[label1,label2]{}
% %% \affiliation[label1]{organization={},
% %%             addressline={},
% %%             city={},
% %%             postcode={},
% %%             state={},
% %%             country={}}
% %%
% %% \affiliation[label2]{organization={},
% %%             addressline={},
% %%             city={},
% %%             postcode={},
% %%             state={},
% %%             country={}}

% \author{} %% Author name

% %% Author affiliation
% \affiliation{organization={},%Department and Organization
%             addressline={}, 
%             city={},
%             postcode={}, 
%             state={},
%             country={}}

%% Authors (linked to affiliation [1])
\author[1]{Mohammad Abu Shaira\corref{cor1}}
\ead{ShairaAbu-Shaira@my.unt.edu}

\author[1]{Yunhe Feng}
\ead{Yunhe.Feng@unt.edu}

\author[1]{Heng Fan}
\ead{Heng.Fan@unt.edu}

\author[1]{Weishi Shi}
\ead{Weishi.Shi@unt.edu}
%% Corresponding author footnote
\cortext[cor1]{Corresponding author.}

%% Affiliation [1]
\affiliation[1]{%
  organization={Department of Computer Science and Engineering, University of North Texas},
  addressline={3940 N Elm St},
  city={Denton},
  postcode={76207},
  state={TX},
  country={USA}
}

%% Abstract
\begin{abstract}
%% Text of abstract
Real-world data sets often exhibit temporal dynamics characterized by evolving data distributions. Disregarding this phenomenon, commonly referred to as concept drift, can significantly diminish a model's predictive accuracy. Furthermore, the presence of hyperparameters in online models exacerbates this issue. These parameters are typically fixed and cannot be dynamically adjusted by the user in response to the evolving data distribution. This paper introduces Online Classification with Weighted Average (\modelname), an adaptive, hyperparameter-free online classification model equipped with an automated optimization mechanism. \modelname operates by blending incoming data streams with an existing base model. This blending is facilitated by an exponentially weighted moving average. Furthermore, an integrated optimization mechanism dynamically detects concept drift, quantifies its magnitude, and adjusts the model based on the observed data stream characteristics. This approach empowers the model to effectively adapt to evolving data distributions within streaming environments. Rigorous empirical evaluation across diverse benchmark datasets shows that \modelname achieves performance comparable to batch models in stationary environments, maintaining accuracy within 1–3\%, and surpasses leading online baselines by 10–25\% under drift, demonstrating its effectiveness in adapting to dynamic data streams.
\end{abstract}

% %%Graphical abstract
% \begin{graphicalabstract}
% %\includegraphics{grabs}
% \end{graphicalabstract}

% %%Research highlights
% \begin{highlights}
% \item Research highlight 1
% \item Research highlight 2
% \end{highlights}

%% Keywords
\begin{keyword}
%% keywords here, in the form: keyword \sep keyword
Online Learning \sep Adaptive Learning \sep Online Classification \sep Concept Drift \sep Hyperparameters Optimization \sep EWMA

%% PACS codes here, in the form: \PACS code \sep code

%% MSC codes here, in the form: \MSC code \sep code
%% or \MSC[2008] code \sep code (2000 is the default)

\end{keyword}

\end{frontmatter}

%% Add \usepackage{lineno} before \begin{document} and uncomment 
%% following line to enable line numbers
%% \linenumbers

%% main text
%%

%#################################################################################
\section{Introduction}
In today’s data-driven landscape, the ability to process and interpret data in real time is increasingly vital. The surge in dynamic data sources such as IoT, social media, and financial transactions demands rapid and immediate insight \cite{hashem2015rise}. However, traditional batch learning falls short due to assumptions of stationary distributions, the need for complete datasets, and the lack of continuous updates. These limitations hinder effectiveness in fast-evolving tasks. For example, fraud detection systems struggle to track shifting fraudulent behaviors, and pre-trained sentiment models quickly become outdated as social media trends change. As a result, there is a growing need for online learning techniques capable of adapting instantly to incoming data, enabling timely and scalable decision-making.

Classification is concerned with analyzing sets of `objects' to verify if they can be meaningfully grouped into a small set of similar classes \cite{gordon1999classification}. Online classification focuses on key attributes such as scalability, adaptability, and low latency by addressing the limitations of the batch approach \cite{hoi2021online}. This includes relaxing assumptions such as the need for complete access to all data for each computation, the assumption of a fixed and unchanging data distribution (\textit{i.i.d.}), and the assumption of no time constraints for the training process.

Online classification faces significant challenges, including concept drift, where data distributions change over time, making it difficult to maintain model accuracy \cite{mehta2017concept}. Additionally, the presence of hyperparameters in online learning exacerbates the issue, as it is impractical to manually tune them upon every occurrence of concept drift. Moreover, the absence of a decay mechanism further worsens the situation, as achieving a balance between adaptability and stability is a critical objective in online learning.

Existing online classification models can be categorized into deep learning-based models and classical machine learning approaches. ODL methods tend to perform well in batch settings because they rely on multi-epoch training, whereas one-pass variants often underperform unless enhanced through architectural or optimization techniques \cite{hu2021one, valkanasmodl}.  In addition, ODL methods face several limitations, including high per-instance computational cost compared to batch training, inefficient hardware usage, and large memory requirements due to storing weights, gradients, and activations \cite{goodfellow2016deep}. Catastrophic forgetting remains a persistent challenge in non-stationary environments \cite{sulaiman2025online}, and deep models typically require much more data to converge than simple ones \cite{sahoo2018online}. Hence, a major focus of this paper is to analyze simple models that typically fall within the category of classical machine learning approaches. 

While classical approaches have historically supported real-time learning, the landscape of online classification algorithms now encompasses a broad spectrum of methods, from simple to advanced, designed for continuous adaptation to streaming data with minimal computational and memory overhead \cite{gama2014survey}. However, despite these advantages, several limitations persist across these algorithms. Some are ineffective at classifying non-linearly separable data \cite{bifet2009adaptive}, while others underperform in classifying borderline instances \cite{crammer2006online}. Additional difficulties arise in managing multi-class classification and imbalanced data streams in real-time contexts \cite{brzezinski2013reacting, krawczyk2016learning}. Many models also fail to adapt effectively to concept drift in real time. This challenge is compounded by the presence of hyperparameters, which are impractical to tune manually with each drift event. Additionally, some algorithms are often sensitive to outdated data, which can impair predictive accuracy due to the absence of a weighting mechanism \cite{bifet2010moa}.

Online hyperparameter-free models are essential due to the impracticality of fine-tuning hyperparameters online after every occurrence of concept drift. To elucidate the concept of hyperparameters within the scope of this research, we clearly distinguish between \textit{hyperparameters} and \textit{configuration parameters}. Configuration parameters are fixed at deployment time and reflect user preferences or operational requirements. For example, choosing a higher false-alarm probability (user choice, $\rho = 0.10$) reflects a stricter approach in classifying subtle changes as drift, whereas selecting a lower probability (user choice, $\rho = 0.01$) reflects a more relaxed approach. Another example is the choice of KPI, some users may prefer accuracy, others may prefer loss, and others may employ multiple KPIs. In contrast, \textit{hyperparameters} are trainable settings that directly influence the learning process and typically require optimization through training and validation, such as learning rate, or regularization strength. In this work, when we describe \modelname as hyperparameter-free, we mean that it does not require such trainable parameters, but instead only relies on user-defined configuration parameters that remain constant during operation and serve the user’s preferences.

% In light of these limitations, this paper proposes a model designed to address three critical challenges in online learning: handling concept drift, eliminating reliance on hyperparameters by adaptive optimization, and incorporating effective weighting or decay mechanisms. This work introduces \modelname, a drift-aware, hyperparameter-free, and decay-based online classifier. Key contributions include: (1) Proposing a novel strategy for dynamically redefining the model in response to emerging data patterns. (2) Featuring a novel proactive in-memory built-in drift detection and adaptation mechanism, which employs a variable threshold technique and uses key performance indicators (KPIs) without assuming any specific data distribution, making it suitable for high-dimensional datasets and large-scale data streams. (3) The method demonstrates a competitive on par performance with batch models in typical classification scenarios, in addition, outperforms leading online methods in dynamic settings by effectively managing concept drift in dynamic data streams, highlighting \modelname's effectiveness and adaptability in real-time learning. The source code and datasets utilized in this study can be accessed 
% through our GitHub repository: 
% \hyperlink{https://github.com/anonymous273800/OLC-WA}{https://github.com/anonymous273800/OLC-WA}.

In light of these limitations, this paper proposes a model designed to address three critical challenges in online learning: handling concept drift, eliminating reliance on hyperparameters by adaptive optimization, and incorporating effective weighting or decay mechanisms. This work introduces \modelname, a drift-aware, hyperparameter-free, and decay-based online classifier. Key contributions include: (1) Proposing a novel strategy for dynamically redefining the model in response to emerging data patterns. (2) Featuring a novel proactive in-memory built-in drift detection and adaptation mechanism, which employs a variable threshold technique and uses key performance indicators (KPIs) without assuming any specific data distribution, making it suitable for high-dimensional datasets and large-scale data streams. (3) Explicitly managing false positives through a user-specified false-alarm probability $\rho$, which guarantees a constant false-positive rate and allows practitioners to control the trade-off between sensitivity and robustness. (4) Enabling seamless conversion of batch classifiers into online models, thereby providing a practical pathway for adapting established methods to streaming environments. (5) The method demonstrates a competitive on par performance with batch models in typical classification scenarios, in addition, outperforms leading online methods in dynamic settings by effectively managing concept drift in dynamic data streams, highlighting \modelname's effectiveness and adaptability in real-time learning.
The source code and datasets utilized in this study can be accessed through our GitHub repository: \hyperlink{https://github.com/anonymous273800/OLC-WA}{https://github.com/anonymous273800/OLC-WA}.

%#################################################################################

%##################################################################

\section{Related Work}
\stepcounter{subsection}
\noindent\textbf{\thesubsection\ Online DL Classifiers:}

Various online deep learning classification models have been proposed to address the challenges of learning from data streams. For instance, Hedge Backpropagation (HBP) \cite{sahoo2018online} leverages a weighted ensemble of classifiers at different depths, allowing dynamic model complexity. Batch-level Distillation \cite{fini2020online} uses a two-stage update scheme to balance plasticity and retention. Other approaches include instance-aware routing \cite{chen2020mitigating}, elastic adaptation via intermediate classifiers \cite{su2024elastic}, and bilevel optimization for drift handling \cite{han2021bilevel}.

Despite recent advancements, online deep learning (ODL) models face persistent challenges. Most rely on multi-epoch training, violating strict online constraints, while one-pass variants typically underperform unless compensated by architectural or optimization tricks \cite{hu2021one, valkanasmodl}. ODL methods are also hindered by high per-instance computation, inefficient hardware use, and large memory footprints due to storage of weights, gradients, and activations \cite{goodfellow2016deep}. Catastrophic forgetting remains a core issue in non-stationary settings \cite{sulaiman2025online}, and deep models often require significantly more data to converge compared to shallow methods \cite{sahoo2018online}.

Recent deep drift-adaptation frameworks extend ODL models by incorporating meta-learning and recurrent mechanisms to handle evolving data distributions. For example, Yu et al.~\cite{yu2022metalearningdrift} proposed a meta-learning approach that rapidly adapts pre-trained deep models to new concepts through episodic optimization, while Suryawanshi et al.~\cite{suryawanshi2023adaptive} introduced an adaptive-window recurrent neural network capable of retaining both short- and long-term dependencies under drift. Other meta-learning and ensemble-based frameworks~\cite{chen2025twostage, yu2024online} also demonstrate strong adaptability in non-stationary environments but typically incur high computational and memory costs, require complex hyperparameter tuning, and depend on multi-epoch updates, limiting their practicality in strict one-pass online learning scenarios.

These challenges, whether unresolved or only partially mitigated, highlight the limitations of current ODL approaches. This motivates a renewed focus on classical online machine learning models, whose simplicity, efficiency, and transparency offer valuable trade-offs in evolving stream environments.
\newline\newline
\begin{table}[htbp]
{\scriptsize % was \footnotesize
\setlength{\tabcolsep}{3pt}   % was 4pt
\renewcommand{\arraystretch}{0.98} % was 1.05
% Optional: tighten caption spacing if using caption pkg
% \captionsetup{skip=2pt}

\begin{tabularx}{\linewidth}{p{1.7cm} X c c c c}
\caption{Feature Comparison of Online Machine Learning Classifiers\label{tab:online-classifiers-features}}\\
\toprule
\textbf{Classifier} & \textbf{Features (\textbf{+}) Limitations (\textbf{--})} &
\rot{\textbf{Drift Aware}} &
\rot{\textbf{Decayed}} &
\rot{\textbf{Hparam Free}} &
\rot{\textbf{Probabilistic}} \\
\midrule

\textbf{PLA}\cite{rosenblatt1958perceptron} &
\textbf{+} Efficient for linearly separable data; \textbf{--} Lacks robustness to noise and non-linear patterns; no probabilistic output\cite{novikoff1962convergence}. &
\xmark & \xmark & \cmark & \xmark \\

\textbf{LMS}\cite{widrow1960adaptive} &
\textbf{+} Minimizes mean squared error; \textbf{--} Highly sensitive to learning rate; poor performance in non-stationary environments\cite{fontenla2013online}. &
\xmark & \xmark & \xmark & \xmark \\

\textbf{OLR}\cite{hazan2014logistic} &
\textbf{+} Provides probabilistic predictions; \textbf{--} Requires careful hyperparameter tuning\cite{murphy2012machine}. &
\xmark & \xmark & \xmark & \cmark \\

\textbf{ONB}\cite{gumus2014online} &
\textbf{+} Computationally efficient; \textbf{--} Independence assumption; vulnerable to concept drift; zero-frequency issues\cite{rennie2003tackling}. &
\xmark & \xmark & \cmark & \cmark \\

\textbf{PA}\cite{crammer2006online} &
\textbf{+} Adapts aggressively to errors; theoretical guarantees; \textbf{--} No probabilistic outputs; relies on fixed hparams. &
\xmark & \xmark & \xmark & \xmark \\

\textbf{VFDT}\cite{domingos2000mining} &
\textbf{+} Scalable for large streams; \textbf{--} Memory-intensive; sensitive to drift; biased under class imbalance\cite{bifet2010leveraging}. &
\xmark & \xmark & \cmark & \xmark \\

\bottomrule
\end{tabularx}
}
\end{table}

\begin{table}[htbp]
{\scriptsize
\setlength{\tabcolsep}{3pt}
\renewcommand{\arraystretch}{0.98}

\begin{tabularx}{\linewidth}{p{2.1cm} X}
\caption{Summary of Concept Drift Detection Techniques\label{tab:related-work-summary-drift-detection-techniques}}\\
\toprule
\textbf{Method} & \textbf{Features (+) and Limitations (--)} \\
\midrule

\textbf{DDM}\cite{gama2004learning} &
\textbf{+} Simple; computationally efficient; two levels: warning and drift. \textbf{--} Limited to classification with abrupt drift; less effective for gradual drifts; effectiveness depends on fixed warning ($k_w = 2$) and drift ($k_d = 3$) multipliers not adapted at runtime; assumes \textit{i.i.d.} Bernoulli errors which can reduce accuracy or cause false alarms. \\

\textbf{EDDM}\cite{baena2006early} &
\textbf{+} Simple; improves gradual drift detection while maintaining performance on abrupt drift. \textbf{--} Limited to classification; performance depends on fixed warning ($\alpha$) and drift ($\beta$) thresholds set before runtime; assumes \textit{i.i.d.} Bernoulli errors which can reduce accuracy or cause false alarms. \\

\textbf{ADWIN}\cite{bifet2007learning} &
\textbf{+} Generic; statistical guarantees. \textbf{--} Detection delays; computationally intensive due to continuous comparison of multiple sub-windows; requires a confidence parameter ($\delta$). \\

\textbf{PL}\cite{bach2008paired} &
\textbf{+} Generic with classification focus. \textbf{--} Higher computation \& memory since two models run and update in parallel; performance depends on reactive window size ($w$) and misclassification threshold ($\theta$). \\

\textbf{KSWin}\cite{raab2020reactive} &
\textbf{+} Non-parametric with no prior data assumptions. \textbf{--} Requires tuning of significance level ($\alpha$) and window size; overhead from maintaining two sliding windows; detection delay; not suited for high-dimensional data as KS test is univariate (each dimension monitored separately); sensitive to window size ($n$), split ratio, and $\alpha$. \\

\textbf{ITA}\cite{dasu2006information} &
\textbf{+} Generic; non-parametric with no prior data assumptions. \textbf{--} Computational cost from multi-dimensional distribution estimation and bootstrapping, limiting high-speed streams; implementation complexity due to robust multi-dimensional density estimation; parameter sensitivity to window size and significance level ($\alpha$). \\

\bottomrule
\end{tabularx}
}
\end{table}

\begin{table}[htbp]
{\scriptsize
\setlength{\tabcolsep}{3pt}
\renewcommand{\arraystretch}{0.98}

\begin{tabularx}{\linewidth}{p{3cm} X}
\caption{Summary of Concept Drift Adaptation Techniques\label{tab:related-work-summary-dift-adaptation-techniques}}\\
\toprule
\textbf{Method} & \textbf{Features (+) and Limitations (--)} \\
\midrule

\makecell[l]{\bfseries Hyperparam \\ \bfseries Optimization} &
\textbf{+} Tuning via grid, random, Bayesian, or gradient-based methods. \textbf{--} No consensus on methods for online learning; batch methods, if used, must revisit the data, violating the one-pass assumption of online learning. \\

\makecell[l]{\bfseries Model \\ \bfseries Replacement\cite{celik2023online}} &
\textbf{+} Generic; suited for abrupt shifts. \textbf{--} Loses past knowledge; not ideal for gradual drift. \\

\makecell[l]{\bfseries Model \\ \bfseries Retraining\cite{baier2021detecting}} &
\textbf{+} Generic; suitable for moderate-latency applications. \textbf{--} High cost; unsuitable for real-time settings. \\

\makecell[l]{\bfseries Ensemble \\ \bfseries Learning\cite{sun2016online, krawczyk2017ensemble}} &
\textbf{+} Generic; combines multiple models for robust predictions. \textbf{--} High memory/inference overhead; requires tuning. \\

\makecell[l]{\bfseries Updating \\ \bfseries Model \\ \bfseries Structure\cite{hulten2001mining, kolter2007dynamic}} &
\textbf{+} Dynamically adapts model (e.g., trees, weights) based on performance, usually tree-based or graph-structured learners. \textbf{--} Implementation complexity; computational cost. \\

\bottomrule
\end{tabularx}
}
\end{table}

\stepcounter{subsection}
\noindent\textbf{\thesubsection\ Online ML Classifiers:}
Classical online classification models present diverse strengths and limitations. The Perceptron (PLA) \cite{rosenblatt1958perceptron}, while efficient for linearly separable data, lacks robustness to noise and non-linear patterns and does not offer probabilistic outputs \cite{novikoff1962convergence}. Widrow-Hoff (LMS) \cite{widrow1960adaptive} minimizes mean squared error but is highly sensitive to the choice of learning rate and struggles in non-stationary environments \cite{fontenla2013online}. Online Logistic Regression (OLR) \cite{hazan2014logistic} provides probabilistic predictions but requires careful hyperparameter tuning \cite{murphy2012machine}. Online Naive Bayes (ONB) \cite{gumus2014online} is computationally efficient yet limited by the independence assumption and vulnerable to concept drift and zero-frequency issues \cite{rennie2003tackling}. Passive-Aggressive (PA) algorithms \cite{crammer2006online} adapt aggressively to errors and offer theoretical guarantees, but lack probabilistic outputs and depend on fixed hyperparameters. Hoeffding Trees (VFDT) \cite{domingos2000mining} are scalable for large streams, though they are memory-intensive, sensitive to drift, and exhibit bias under class imbalance \cite{bifet2010leveraging}.
\newline\newline
\stepcounter{subsection}
\noindent\textbf{\thesubsection\ Concept Drift Management:}
Concept drift in online learning is commonly managed through detection and adaptation strategies, which are typically treated as separate components. Prominent CD detection techniques include DDM \cite{gama2004learning}, EDDM \cite{baena2006early}, ADWIN \cite{bifet2007learning}, PL \cite{bach2008paired}, KSWin \cite{raab2020reactive}, and ITA \cite{dasu2006information}. These methods are reactive, identifying drift only after it has occurred and its impact on model performance becomes apparent. Many rely on detecting distributional shifts, as opposed to monitoring performance-based indicators, which are often sensitive to noise and data variance especially in high dimensional data. Additionally, they generally lack the ability to distinguish between different types of drift, limiting their versatility across varying scenarios. For adaptation, hyperparameter optimization is commonly employed; however, there is no established consensus on how to effectively optimize hyperparameters in online learning \cite{barbaro2018tuning}. Traditional batch-based methods, such as grid search and random search, require multiple passes over the data \cite{gama2014indre}, thereby violating the single-pass constraint inherent to online learning.  Additional strategies include model replacement \cite{celik2023online}, which discards prior knowledge; model retraining \cite{baier2021detecting}, which incurs high computational cost; ensemble learning \cite{sun2016online, krawczyk2017ensemble}, which introduces memory and inference overhead; and structural model updates \cite{hulten2001mining, kolter2007dynamic}, which are often complex to implement and maintain.

% In summary, while deep learning models have demonstrated impressive capabilities in batch settings, they often show fundamental unsuitability for online learning due to their reliance on multiple epochs, high memory consumption, slow convergence, and vulnerability to catastrophic forgetting \cite{hu2021one, goodfellow2016deep, sulaiman2025online}. In contrast, classical online classification models are lightweight and update-efficient, but still face several persistent limitations. Many lack decay or weighting mechanisms that are essential for balancing adaptability and stability in dynamic environments \cite{hayes2022online, liu5136906online, titsiaskalman, gama2013evaluating, haug2022towards}. Several do not produce probabilistic outputs, limiting their applicability in confidence-aware decision-making \cite{crammer2006online, freund1998large, domingos2000mining, gama2014survey, bottou2003stochastic}. They are also sensitive to noise and outliers \cite{khardon2007noise, panup2021stochastic}, rely on restrictive assumptions such as feature independence in Naive Bayes \cite{gumus2014online}, and are prone to bias in imbalanced scenarios \cite{domingos2000mining}. A widespread challenge is their limited adaptability to concept drift \cite{gama2014survey, kumar2015survey, lu2018learning, galmeanu2021concept}, further exacerbated by static hyperparameters that cannot dynamically adjust to evolving data streams \cite{rosenblatt1958perceptron, novikoff1962convergence, widrow1960adaptive, hazan2014logistic, murphy2012machine, crammer2006online}. 

In summary, this literature synthesis summarized in Tables \ref{tab:online-classifiers-features}, \ref{tab:related-work-summary-drift-detection-techniques} and \ref{tab:related-work-summary-dift-adaptation-techniques} underscores the need for a streamlined classical online machine learning approach that is unaffected by the limitations of deep learning in online settings, capitalizes on the strengths of classical methods, and integrates essential mechanisms to fill existing gaps. Our proposed approach aims to 1) bridge the trade-off between flexibility and robustness by explicitly addressing concept drift, 2) eliminating hyperparameters that are impractical to retune during deployment on every occurrence of concept drift, 3) incorporating decay or weighting mechanisms for adaptive stability, and 4) providing probabilistic outputs to enable confidence-aware predictions, thereby fulfilling key deficiencies in current online classifiers.

% \subsection{Weighting Strategies for Online Learning}

% Various weighting strategies enhance online learning adaptability by prioritizing recent data and down-weighting older observations. EWMA \cite{holt2004forecasting} applies exponential decay for smooth drift adaptation, while Sliding Windows \cite{kuncheva2009window, deypir2012towards} retain only the latest data points. LWMA \cite{makridakis2008forecasting, hyndman2018forecasting} improves trend sensitivity through linearly increasing weights. More adaptive approaches include reinforcement learning-based weighting \cite{xiang2023concept, yahmed2023intentional}, which optimizes forgetting rates via feedback, and kernel-based mechanisms \cite{van2012kernel, maldonado2021time, csato2002sparse, berntorp2021online}, which incorporate temporal decay into kernels or posterior updates—particularly effective in non-linear, non-stationary settings \cite{kivinen2004online}.

%##############################################################
\section{Method}

\subsection{Problem Settings}

Let \( \mathcal{X} \subseteq \mathbb{R}^d \) denote the input feature space and \( \mathcal{Y} = \{1, 2, \dots, K\} \) represent the set of possible class labels. At each time step \( t \), an instance \( (\mathbf{x}_t, y_t) \) arrives sequentially, where \( \mathbf{x}_t \in \mathcal{X} \) is a feature vector and \( y_t \in \mathcal{Y} \) is the true class label. The data stream is presumed to be infinitely long and must be handled in a sequential manner, with only a limited amount of data stored in memory at any point in time \cite{gama2014survey}. The objective of our online classification model is to update the decision function \( f_t: \mathcal{X} \to \mathcal{Y} \) incrementally with the goal of minimizing the cumulative classification loss over time. Typically, data points in an online learning scenario are processed in small increments, allowing models to update iteratively as new information arrives. However, the data generation process is subject to changes over time, leading to a phenomenon known as \textit{concept drift}, which refers to unexpected shifts in the underlying data distribution. Formally, consider a data stream \( \{(\mathbf{x}_1, y_1), (\mathbf{x}_2, y_2), \dots\} \) generated from an unknown joint probability density function \( p(\mathbf{x}, y) \) that represents the concept to be learned. Concept drift can be expressed as: \(
\boldsymbol{\exists \mathbf{x}, y \; : \; p_{t_n}(\mathbf{x}, y) \neq p_{t_{n+1}}(\mathbf{x}, y)}
\), 
where \( t_n \) and \( t_{n+1} \) denote two consecutive time points, illustrating the evolving nature of the data. Concept drift can manifest in various forms, including abrupt, incremental, and gradual changes, each presenting unique challenges to the learning system \cite{webb2016characterizing,gama2014survey}.

\subsection{Learning Objective}

The \textbf{learning objective} governs the update of the model's weight vector, $\mathbf{w}$, at each step $t$. It is formulated as an optimization problem that explicitly balances \textbf{stability} (retaining prior knowledge) and \textbf{adaptability} (incorporating new information). The weight vector for the next step, $\mathbf{w}_{t+1}$, is determined by minimizing the following objective function:

\begin{multline}
\mathbf{w}_{t+1} = \underset{\mathbf{w}}{\arg\min} \Bigg(
\underbrace{\frac{1}{2} \| \mathbf{w} - \mathbf{w}_t \|^2}_{\text{Stability Regularizer}} +
\underbrace{\alpha \mathcal{L}(y_{\text{inc}}, f(\mathbf{w}; \mathbf{X}_{\text{inc}}))}_{\text{Adaptability Loss}} \\[-0.5cm]
+ \underbrace{(1 - \alpha) \mathcal{L}(y_{\text{base}}, f(\mathbf{w}; \mathbf{X}_{\text{base}}))}_{\text{Retention Loss}}
\Bigg)
\end{multline}

The objective function comprises three distinct terms, each contributing uniquely to the online learning process:

\begin{enumerate}[labelsep=0.8em]
    \item \textit{Stability Regularizer:} 
    The first term, $ \frac{1}{2} \| \mathbf{w} - \mathbf{w}_t \|^2 $, acts as an \textbf{$L_2$ regularization} penalty. It promotes \textbf{smooth weight updates} by penalizing significant deviations from the previous weight vector $\mathbf{w}_t$. This encourages the model to retain its established knowledge and mitigates the risk of catastrophic forgetting when presented with new data.

    \item \textit{Adaptability Loss:}
    The second term, $ \alpha \mathcal{L}(y_{\text{inc}}, f(\mathbf{w}; \mathbf{X}_{\text{inc}})) $, is the \textbf{logistic loss} computed over the \textbf{incremental mini-batch} of new data, $(\mathbf{X}_{\text{inc}}, y_{\text{inc}})$. The weighting factor $ \alpha \in (0,1] $ emphasizes recent observations, enhancing the model's \textbf{adaptability} to evolving data patterns. A higher value of $\alpha$ allows the model to more rapidly incorporate the information from the current mini-batch.

    \item \textit{Retention Loss:}
    Conversely, the third term, $ (1 - \alpha) \mathcal{L}(y_{\text{base}}, f(\mathbf{w}; \mathbf{X}_{\text{base}})) $, represents the \textbf{logistic loss} associated with a representative \textbf{base (historical) mini-batch}, $(\mathbf{X}_{\text{base}}, y_{\text{base}})$. By minimizing this term, the model is compelled to maintain a satisfactory performance on past observations, thereby ensuring the \textbf{retention} of previously learned knowledge through the weight $ (1 - \alpha) $.
    
\end{enumerate}

\subsection{Online Learning Parameter Update}

\subsubsection{General Overview}
The \modelname model incrementally learns from streaming data by maintaining two decision boundaries: the \textbf{base boundary} \( f_{\text{base}}(x) \), representing accumulated past knowledge, and the \textbf{incremental boundary} \( f_{\text{inc}}(x) \), derived from the most recent mini-batch at time \( t \), where \( x \) is the feature vector. 
Initially, the first available mini-batch is used to establish the base boundary \( f_{\text{base}}(x_0) \), from which the corresponding weight vector \( W_{\text{base}} \) is obtained. 
As new data arrive, each incoming mini-batch defines an incremental boundary \( f_{\text{inc}}(x_t) \), parameterized by the weight vector \( W_{\text{inc}} \).

Their respective norm vectors, \( V_{\text{base}} \) and \( V_{\text{inc}} \), are combined using the Exponentially Weighted Moving Average (EWMA). The smoothing factor \( \alpha \in (0,1] \) emphasizes recent data via \( \alpha \), while \( 1 - \alpha \) preserves historical knowledge. The resulting average norm vector \( V_{\text{Avg}} \) is updated iteratively using the intersection point \( P_{\text{int}} \) of the two boundaries to redefine the new base model \( W_{\text{base}} \), progressively aligning it with \( W_{\text{inc}} \) as new observations arrive.

\subsubsection{Construction of Base and Incremental Hyperplanes}

As an online learning model, \modelname processes a data stream sequentially, updating its decision boundary with each incoming mini-batch. At any time $t$, the model maintains two representations: a base parameter vector ($\mathbf{W}_{\text{base}}$) derived from past data and an incremental parameter vector ($\mathbf{W}_{\text{inc}}$) constructed from the most recent mini-batch $\{\mathbf{X}_{\text{inc}}, \mathbf{y}_{\text{inc}}\}$. Both $\mathbf{W}_{\text{base}}$ and $\mathbf{W}_{\text{inc}}$ implicitly define hyperplanes and are obtained by minimizing the logistic loss ($\mathcal{L}_{\text{NLL}}$) using a built-in convex optimization solver.

The decision boundary is updated incrementally over time. The process begins at time $t_0$ when the initial mini-batch $\{\mathbf{X}_{\text{base}}, \mathbf{y}_{\text{base}}\}$ is processed to construct the initial $\mathbf{W}_{\text{base}}(t_0)$. Subsequently, at time $t_1$, the incoming mini-batch $\{\mathbf{X}_{\text{inc}}, \mathbf{y}_{\text{inc}}\}$ is used to construct $\mathbf{W}_{\text{inc}}(t_1)$. The model then combines the corresponding base normal vector $\mathbf{V}_{\text{base}}$ and incremental normal vector $\mathbf{V}_{\text{inc}}$ using an Exponentially Weighted Moving Average (EWMA), weighted by $\alpha$, to compute the averaged vector $\mathbf{V}_{\text{avg}}$. This $\mathbf{V}_{\text{avg}}$, along with the calculated intersection point $\mathbf{P}_{\text{int}}$, defines the updated base parameter vector, $\mathbf{W}_{\text{base}}(t_1)$, for time $t_1$.

\subsubsection{Norm Vectors and Normalization}

The decision hyperplane for both the base and incremental models is defined by the function $f(\mathbf{x}) = \mathbf{V} \cdot \mathbf{x} + b$. The \textbf{norm vector} ($\mathbf{V}$) is extracted directly from the feature weights of the corresponding full parameter vector, $\mathbf{W}$ (which includes the bias term $b$). Specifically, $\mathbf{V}_{\text{base}} = \mathbf{W}_{\text{base}}[1:d]$ and $\mathbf{V}_{\text{inc}} = \mathbf{W}_{\text{inc}}[1:d]$. The norm vector $\mathbf{V}$ fundamentally defines the hyperplane's geometry. Its \textbf{magnitude}, computed as the Euclidean length $\|\mathbf{V}\| = \sqrt{\sum_{i=1}^{d} V_i^2}$, is inversely related to the distance between the decision boundary and the nearest data points. The vector's \textbf{direction} ($\hat{\mathbf{V}}$), defined by the unit vector $\hat{\mathbf{V}} = \frac{\mathbf{V}}{\|\mathbf{V}\|}$, establishes the hyperplane's precise orientation in the feature space.

Normalization is a critical step prior to the Exponentially Weighted Moving Average (EWMA) update to ensure that the combination of $\mathbf{V}_{\text{base}}$ and $\mathbf{V}_{\text{inc}}$ reflects a pure blend of \textbf{orientations} rather than being distorted by disparate scales. The norm vectors are normalized to unit length, yielding $\hat{\mathbf{V}}_{\text{base}}$ and $\hat{\mathbf{V}}_{\text{inc}}$. The subsequent EWMA step, $\mathbf{V}_{\text{avg}} = (1 - \alpha)\hat{\mathbf{V}}_{\text{base}} + \alpha\hat{\mathbf{V}}_{\text{inc}}$, thus guarantees that the weighted average is a \textbf{convex combination of directions}. Utilizing un-normalized vectors would cause the direction of $\mathbf{V}_{\text{avg}}$ to be unduly dominated by the vector with the larger magnitude, which would undermine the calculated adaptive weighting factor $\alpha$. Normalization ensures that $\alpha$ truly represents the desired proportional influence of the new incremental information.

\subsubsection{The Weighted Average Norm Vector}
The weighted average norm vector $\mathbf{V}_{\text{avg}}$ defines the orientation of the final decision boundary and is constructed by combining the normalized base and incremental norm vectors ($\mathbf{V}_{\text{base}}$ and $\mathbf{V}_{\text{inc}}$) using a weighted average (EWMA). This process employs the weighting factor $\alpha$, which is crucial for balancing the model's stability (retention of $\mathbf{V}_{\text{base}}$) against its adaptability (incorporation of $\mathbf{V}_{\text{inc}}$). A fixed $\alpha$ is recognized to provide only a static trade-off that is insufficient for dynamic data streams; however, the \modelname framework overcomes this limitation by dynamically adjusting $\alpha$. The general construction is given by Equation~\ref{eq:olc-wa-gen-equation}.

\modelname uses EWMA \cite{shumway2000time} to calculate the weighted average of the base and incremental decision boundaries. EWMA is chosen among weighting strategies for its efficient, smooth adaptation to concept drift while preserving past knowledge with exponentially decaying influence. Its lightweight \( O(d) \) complexity ensures computational efficiency, making it ideal for real-time online learning.

The weighted average formula, detailed in Equation $\ref{eq:olc-wa-gen-equation}$, maintains a consistent structure regardless of the data's dimensionality, combining the norm vectors of the base and incremental decision hyperplanes using a dynamically computed weight $\alpha$ and its complement $1 - \alpha$. This factor $\alpha$ is crucial as it controls the influence of recent versus historical data. By default, both components are initialized with equal weights ($\alpha=0.5$), unless a user-specified initial weight is provided. Critically, these weights are automatically computed and dynamically adjusted per iteration to either accelerate adaptation or enhance robustness against concept drifts. This automated adjustment highlights \modelname as a hyperparameter-free model, as it removes the need for manual tuning of the $\alpha$ parameter. 

This process is general across all drift types; what differentiates it for a specific detected drift is the smoothing factor \(\alpha\). The factor \(\alpha \in (0,1]\) controls the balance between adaptability and stability, where higher values emphasize recent data (favoring adaptation), while lower values preserve historical knowledge (favoring stability).
The methodology for the automated assignment of $\alpha$ is detailed later in this paper.

\begin{equation}
\label{eq:olc-wa-gen-equation}
    V_{\text{Avg}} = {(1 - \alpha) \cdot V_{\text{base}} + \alpha \cdot V_{\text{inc}}}
\end{equation}

\subsubsection{Intersection Point}

The \textbf{intersection} of the base decision boundary, $f_{\text{base}}(\mathbf{x})$, 
and the incremental decision boundary, $f_{\text{inc}}(\mathbf{x})$, 
is determined by solving the resulting \textit{linear system of equations}. This solution is typically obtained using \textit{standard numerical techniques} such as 
\textit{Gaussian elimination} or equivalent \textit{linear system solvers} available in computational libraries. 

The core procedure for finding the intersection of the base and incremental hyperplanes remains the same regardless of the space's dimensionality: it involves solving a linear system comprising two equations (one for each hyperplane) in $N$ variables. However, the nature of the resulting intersection changes with dimensionality. In the familiar two-dimensional (2D) plane ($N=2$), solving two equations with two variables typically yields a single, exact point of intersection. As dimensionality increases (e.g., $N=3$ or higher), solving two equations in $N$ variables results in a solution set with $N-2$ degrees of freedom. This means the intersection is no longer a single point, but an entire subspace, specifically an $(N-2)$-dimensional flat, such as a line in 3D space. While methods like Gaussian elimination can systematically find this solution set, for the purpose of defining the new decision boundary, we only require any single point ($\mathbf{P}_{\text{int}}$) that lies on this intersection subspace, which is sufficient to anchor the combined hyperplane.

The geometric definition of the updated base hyperplane is predicated on the existence of an \textit{intersection point} between the base and incremental hyperplanes. However, this critical geometric mechanism is compromised when hyperplanes are \textit{parallel} or \textit{coincident}, resulting in no intersection. The algorithmic handling of these exceptional cases is discussed thoroughly. Furthermore, this discussion will examine the challenges and potential solutions for extending the geometric intersection mechanism to \textit{non-linear decision boundaries}. The comprehensive analysis of these algorithmic necessities and the robustness of the intersection-based update are detailed under the heading \textit{Geometric Robustness and Hyperplane Intersection} in the Discussion section.

%$$$$$$$$$$$$$$$$$$

\subsubsection{Defining Decision Boundary}

In the realm of N-dimensional geometry, the definition of a new hyper-plane necessitates the utilization of a norm vector and a point \cite{stewart2020calculus}. Let \(\mathbf{n} = \langle n_1, n_2, \ldots, n_N \rangle\) denote the norm vector, representing the directional characteristics of the desired hyper-plane, and let \(\mathbf{P} = (X_{1_p}, X_{2_p}, \ldots, X_{N_p})\) be a point lying within the hyper-plane. To establish the equation of the new hyper-plane, consider an arbitrary point \(\mathbf{Q} = (x_1, x_2, \ldots, x_N)\) on the hyper-plane. The vector connecting \(\mathbf{P}\) and \(\mathbf{Q}\), denoted as \(\overrightarrow{\text{PQ}} = \langle x_1 - X_{1_p}, x_2 - X_{2_p}, \ldots, x_N - X_{N_p} \rangle\), lies within the hyper-plane and is orthogonal to the norm vector \(\mathbf{n}\). Hence, the orthogonality condition can be expressed as \(\mathbf{n} \cdot \overrightarrow{\text{PQ}} = n_1(x_1 - X_{1_p}) + n_2(x_2 - X_{2_p}) + \ldots + n_N(x_N - X_{N_p}) = 0\). This methodology enables the precise definition of a new hyper-plane in N-dimensional space based on its norm vector and a known point lying within the hyper-plane.

\begin{figure}[t]
\centering
\includegraphics[width=\linewidth]{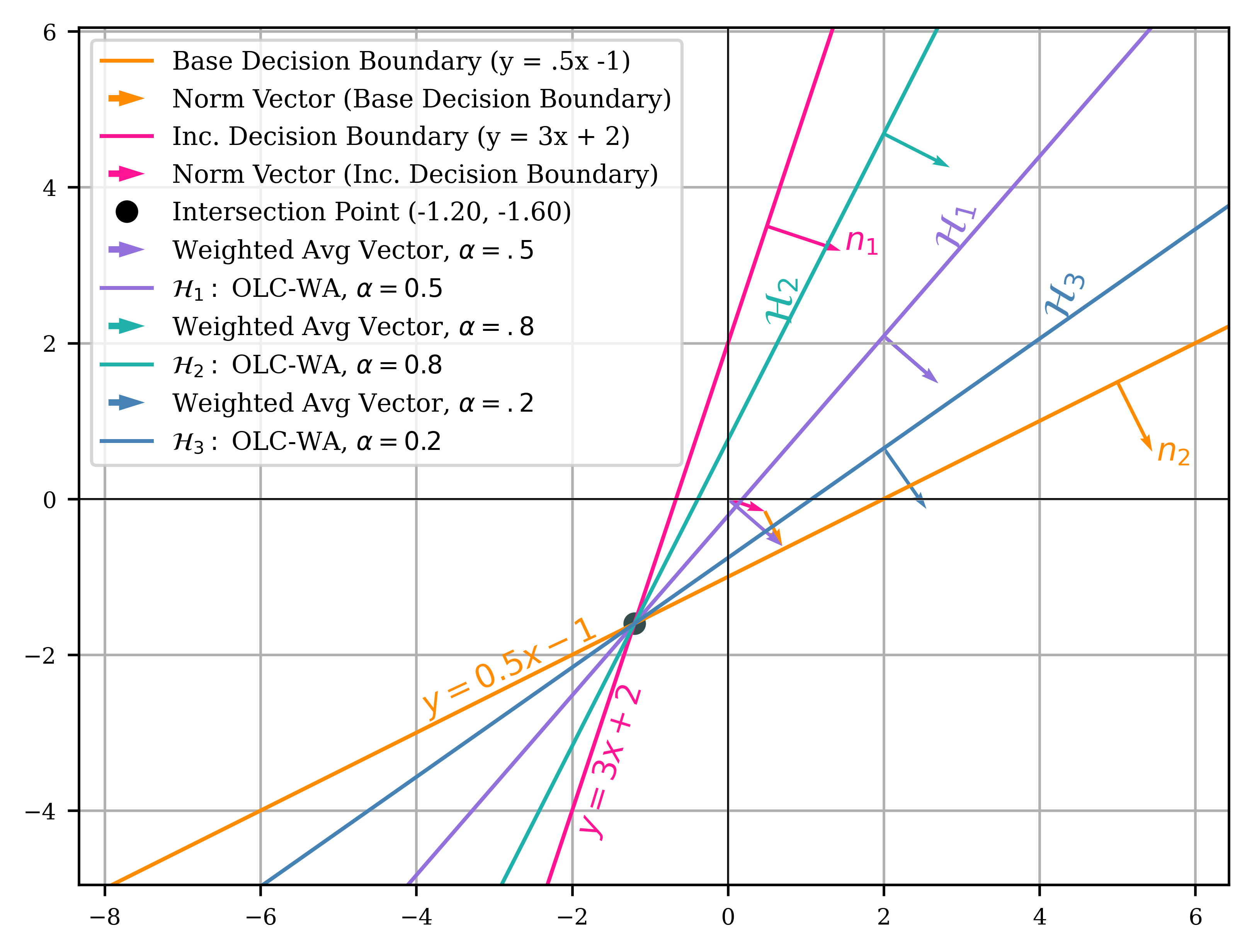}
\caption{\modelname One Step Online Learning}
\label{fig:one_step_update}
\end{figure}

The \modelname model directly applies this geometric principle by utilizing the computed weighted average norm vector $\mathbf{V}_{\text{avg}}$ along with the intersection point ($\mathbf{P}_{\text{int}}$) to define the updated decision boundary for the current time $t$. Figure \ref{fig:one_step_update} provides a comprehensive visualization of the process of defining the updated decision boundary within a two-dimensional space, which subsequently serves as the new base model for the upcoming iteration. This scenario represents an adversarial setting where the incoming data deviates significantly from the base model. The base decision boundary, depicted in dark orange, represents the accumulated knowledge up to time \( (t - 1) \), while the current incoming batch, captured at time \( (t) \), is shown in deep pink. Their respective norm vectors are computed and illustrated using the same color scheme. Assuming \( \alpha = 0.5 \), the weighted average vector, shown in purple, is computed using Equation \ref{eq:olc-wa-gen-equation}, and the new decision boundary, denoted as hyperplane \( \mathcal{H}_1 \), is determined by the intersection of the base and incremental boundaries along with the weighted average norm vector. The figure further illustrates two additional decision boundaries: \( \mathcal{H}_2 \), which corresponds to a higher \( \alpha \) value, enhancing adaptation by prioritizing new data, and \( \mathcal{H}_3 \), which represents a confidence-based approach that emphasizes stability by focusing on reliable data points. These boundaries demonstrate the flexibility of \modelname in handling various scenarios, including adversarial settings where a high \( \alpha \) value facilitates greater adaptation to incoming mini-batches that significantly deviate from the current model, and confidence-based settings where the model adopts a conservative updating strategy by prioritizing established, high-confidence data to ensure stability in dynamic environments.

The \modelname model directly applies this geometric formulation dynamically at each iteration. The updated base hyperplane ($\mathcal{H}_{\text{base}}$) is obtained from the intersection of the previous base and current incremental boundaries, with its orientation determined by the averaged norm vector $\mathbf{V}_{\text{avg}}$ and its location anchored at the intersection point $\mathbf{P}_{\text{int}}$. This geometric definition holds for any dimensionality $N$. Thus, the same mechanism is applied whether the feature space is 2D, 3D, or high-dimensional. Geometrically, this mechanism ensures that the decision boundary does not shift abruptly; instead, it rotates or translates gradually toward the new concept, maintaining crucial continuity and stability between the old and new concepts based on the severity of the drift.

%########################

\subsection{Multi-class classification}

The proposed $\text{\modelname}$ framework is naturally extended to multi-class scenarios by adopting the standard \textbf{One-vs-Rest (OvR)} decomposition strategy. For a problem involving $K$ distinct classes, the system constructs $K$ independent binary classifiers, $\{C_1, C_2, \ldots, C_K\}$. Each classifier $C_k$ is exclusively trained to distinguish its target class ($y=k$) from all others ($y \neq k$). This approach transforms the multi-class streaming problem into $K$ parallel binary streams, where each learner processes the incoming data stream with relabeled outputs specific to its binary task. This design ensures that the model preserves its fundamental adaptive and incremental learning properties across all $K$ decision boundaries simultaneously.

The critical advantage of this OvR extension lies in the \textbf{independent and autonomous adaptation} of each classifier. At every iteration, each classifier $C_k$ independently applies the core $\text{\modelname}$ mechanism: it computes an incremental norm vector ($\mathbf{V}_{\text{inc}}^{(k)}$) from the latest mini-batch, uses the drift detection module to determine a class-specific smoothing factor ($\alpha^{(k)}$), and updates its parameters via the Exponentially Weighted Moving Average (EWMA) to produce $\mathbf{V}_{\text{avg}}^{(k)}$. The base and incremental parameter vectors for each class are combined at every iteration through weighted averaging of their norm vectors ($\mathbf{V}_{\text{avg}}$) and an intersection-based hyperplane update to define the new base parameter vector $\mathbf{W}_{\text{base}}$ as shown in Equation (\ref{eq:update-W-base}). This parallelized adaptation allows each boundary to evolve independently based on the data associated with its respective class, ensuring the model’s overall stability and responsiveness in real time.

During the prediction phase, an input instance $\mathbf{x}$ is evaluated by all $K$ independent classifiers. Each classifier $C_k$ outputs a probability $P(y=k \mid \mathbf{x})$, and the final classification output $\hat{y}$ is determined by selecting the class corresponding to the maximum probability, as defined in Equation (\ref{eq:multiclass-prediction}). This architecture ensures the efficient integration of $K$ continuously refined, EWMA-based binary learners, making $\text{\modelname}$ highly suitable for adaptive multi-class analysis of online data streams.

\begin{equation}
\label{eq:multiclass-prediction}
\hat{y} = \arg\max_{k \in \{1, \ldots, K\}} P(y=k \mid \mathbf{x})
\end{equation}

\begin{equation}
\label{eq:update-W-base}
\mathbf{W}_{\text{base}} \gets \begin{bmatrix}\mathbf{V}_{\text{avg}} \\ -\mathbf{V}_{\text{avg}}^{\!\top}\mathbf{P}_{\text{int}}\end{bmatrix}
\end{equation}

%########################

\subsection{Memory-Based Online Hyperparameter Tuning}
\modelname features a built-in hyperparameter tuning mechanism to compute the smoothing factor \(\alpha\). This approach leverages historical Key Performance Indicators (KPIs), such as cost and accuracy, by organizing them into a normal distribution. Through this process, `low' and `high' limits are determined to evaluate performance deviations of the new decision boundary and quantify the drift. Subsequently, an optimized value of \(\alpha\) is inferred (\(\alpha'\)).

In \modelname, the KPI design extends well beyond accuracy and cost to provide a \textit{flexible and extensible} performance evaluation framework. Users are not limited to predefined metrics; they can integrate a broad range of additional indicators such as \textit{precision, recall, F1-score, AUC, calibration error}, or even \textit{system-level measures} like latency, throughput, or computational efficiency. The framework supports any performance measure, enabling users to construct \textit{customized or composite KPIs} that align precisely with their application-specific goals and operational priorities.

KPIs (e.g., accuracy, loss, cost, or F1-score) are continuously monitored during the online learning process, with each value at time \(t\) stored in the rolling KPI-Window (KPI-Win). For each incoming mini-batch, the model generates predictions \(\hat{y}\), and the selected KPI is computed directly. Specifically, the accuracy is given by \(\text{Accuracy} = \frac{1}{N} \sum_{i=1}^{N} \mathbb{I}(\hat{y}_i = y_i)\), while the logistic loss (cross-entropy) is computed as \(\mathcal{L} = -\frac{1}{N} \sum_{i=1}^{N} [y_i \log(\hat{p}_i) + (1 - y_i)\log(1 - \hat{p}_i)]\). These metrics are updated online and stored in KPI-Win for continuous performance tracking and drift analysis.

The optimization process records KPIs at each time \(t\) in a bounded KPI-Win of size \textit{KWS}, defined in Equation~\ref{eq:window-size} using recent \(N\) points in the rolling window and mini-batch size \(K\). The scaling factor \(\gamma = 0.05\) adjusts \textit{KWS}, constrained between lower (LB) and upper (UB) bounds. An experimental size of 31 is used, though smaller values suit frequent abrupt drift.
\begin{equation}
\label{eq:window-size}
\text{LB} \leq \text{KWS} = \left(\frac{N}{K}\right) \times \gamma \leq \text{UB}    
\end{equation}

In an illustrative example with a KPI window of size 31, the first 30 entries correspond to the historical \textit{Baseline Statistics}, while the 31st entry represents the KPI computed using the current decision boundary (incremental model). Assuming the \textit{Baseline Statistics} follow a normal distribution, a variable threshold $\tau$ can be derived instead of relying on a fixed cutoff, which may not generalize well across different KPIs or data streams with varying characteristics. The threshold is computed as $\tau = z \times \sigma$, where $\sigma$ is the standard deviation and $z$ is the multiplier indicating the number of standard deviations from the mean. Traditionally, $z$ is chosen by the user (e.g., 1.5–2.5 are common values). In our approach to mitigate false positives, however, $z$ is calibrated from a user-specified false-alarm probability $\rho$ via $z = \Phi^{-1}(1-\rho)$. This ensures that the probability of a benign point exceeding the threshold equals $\rho$. Smaller values of $z$ (corresponding to higher $\rho$) make the detector more sensitive to subtle changes, while larger values of $z$ (corresponding to lower $\rho$) yield a more relaxed setting that avoids classifying small fluctuations as drift.

\begin{sidewaystable}[!htbp]
\centering
\captionsetup{justification=centering}
\caption{Concept Drift Dataset Properties.\\
{\scriptsize
Drift Locations: Abrupt (A) marks sudden drift onset. Incremental (I) drift recurs every $K$ points.
Gradual (G) shows drift as a sequence of concepts $C_1$ and $C_2$.
C-Start: Start Concept. C-End: End Concept.
E.D.: Euclidean distance between coefficients in space.
D.M.: Drift Magnitude. D.M.\,(Consec.): D.M.\ between each two consecutive concepts.
}}
\scriptsize
\setlength{\tabcolsep}{4pt}
\renewcommand{\arraystretch}{1.25}
\resizebox{\textheight}{!}{% use textheight instead of textwidth since rotated
\begin{tabular}{l c c c c c c c c c c c c}
\toprule
\textbf{Dataset} &
\makecell{\textbf{Drift}\\\textbf{Type}} &
\textbf{Points} &
\makecell{\textbf{Dimen-}\\\textbf{sions}} &
\textbf{Classes} &
\makecell{\textbf{Drift}\\\textbf{Locations}} &
\makecell{\textbf{Inst./}\\\textbf{Class}} &
\makecell{\textbf{C-Start}\\\textbf{E.D.}} &
\makecell{\textbf{C-End}\\\textbf{E.D.}} &
\makecell{\textbf{C-Start}\\\textbf{SNR}} &
\makecell{\textbf{C-End}\\\textbf{SNR}} &
\makecell{\textbf{D.M.}\\\textbf{(Start→End)}} &
\makecell{\textbf{D.M.}\\\textbf{(Consec.)}} \\
\midrule
\textbf{DS15} & A & 1k & 2 & 2 & 0.5k & 0.5k & 4.24 & 6.90 & 4.48 & 7.28 & 15.52 & 15.52 \\
\textbf{DS16} & A & 10k & 20 & 2 & 5k & 5k & 6.72 & 8.90 & 6.71 & 8.90 & 45.84 & 45.84 \\
\textbf{DS17} & A & 1.5k & 2 & 3 & 0.75k & 0.5k &
\makecell[l]{C0:C1=4.24\\C0:C2=3.53\\C1:C2=5.11} &
\makecell[l]{C0:C1=6.93\\C0:C2=4.99\\C1:C2=4.85} &
13.59 & 16.98 & 16.12 & 16.12 \\
\textbf{DS18} & A & 15k & 20 & 3 & 7.5k & 5k &
\makecell[l]{C0:C1=11.19\\C0:C2=20.14\\C1:C2=8.95} &
\makecell[l]{C0:C1=17.88\\C0:C2=35.78\\C1:C2=17.89} &
40.26 & 71.63 & 52.15 & 52.15 \\
\textbf{DS19} & I & 2.8k & 2 & 2 & every 0.2k [0.6k–2.2k] & 1.4k & 4.12 & 5.51 & 4.32 & 5.36 & 43.08 & $4.78 \pm 1.522$ \\
\textbf{DS20} & I & 21k & 20 & 2 & every 1.5k [4.5k–16.5k] & 10.5k & 13.36 & 13.43 & 13.37 & 13.28 & 120.74 & $13.41 \pm 0.029$ \\
\textbf{DS21} & I & 4.2k & 2 & 3 & every 300 [900–3300] & 1.4k &
\makecell[l]{C0:C1=4.02\\C0:C2=1.92\\C1:C2=2.09} &
\makecell[l]{C0:C1=5.88\\C0:C2=2.69\\C1:C2=3.19} &
8.57 & 10.56 & 40.36 & $4.48 \pm 0.474$ \\
\textbf{DS22} & I & 42k & 20 & 3 & every 3k [9k–33k] & 14k &
\makecell[l]{C0:C1=13.38\\C0:C2=26.82\\C1:C2=13.44} &
\makecell[l]{C0:C1=13.45\\C0:C2=26.84\\C1:C2=13.39} &
53.50 & 53.76 & 241.49 & $26.83 \pm 0.014$ \\
\textbf{DS23} & G & 2.6k & 2 & 2 &
\makecell[l]{[0.8k–C1, 0.2k–C2, 0.2k–C1,\\0.4k–C2, 0.2k–C1, 0.8k–C2]} &
1.3k & 2.71 & 2.74 & 2.91 & 2.74 & 21.21 & $21.28 \pm 0.081$ \\
\textbf{DS24} & G & 26k & 20 & 2 &
\makecell[l]{[8k–C1, 2k–C2, 2k–C1,\\4k–C2, 2k–C1, 8k–C2]} &
13k & 13.36 & 13.40 & 13.42 & 13.37 & 67.08 & $67.09 \pm 0.028$ \\
\textbf{DS25} & G & 3.9k & 2 & 3 &
\makecell[l]{[1.2k–C1, 0.3k–C2, 0.3k–C1,\\0.6k–C2, 0.3k–C1, 1.2k–C2]} &
1.3k &
\makecell[l]{C0:C1=4.12\\C0:C2=8.50\\C1:C2=4.38} &
\makecell[l]{C0:C1=2.77\\C0:C2=7.09\\C1:C2=4.32} &
16.42 & 14.36 & 20.31 & $20.21 \pm 0.036$ \\
\textbf{DS26} & G & 39k & 20 & 3 &
\makecell[l]{[12k–C1, 3k–C2, 3k–C1,\\6k–C2, 3k–C1, 12k–C2]} &
13k &
\makecell[l]{C0:C1=13.39\\C0:C2=26.82\\C1:C2=13.43} &
\makecell[l]{C0:C1=13.44\\C0:C2=26.82\\C1:C2=13.38} &
53.64 & 53.70 & 67.08 & $67.06 \pm 0.009$ \\
\bottomrule
\end{tabular}}
\label{tab:datasets-properties_drift}
\end{sidewaystable}

The monitoring `high' limit is used for KPIs where lower values are favorable (e.g., loss), corresponding to the right tail of the distribution. Conversely, the `low' limit applies to KPIs where higher values are desirable (e.g., accuracy), corresponding to the left tail. For instance, if KPI = Accuracy, a drop into the left tail indicates performance degradation, which in turn triggers the optimization process.

A dynamic, on-the-fly scaled dictionary is constructed to determine the fine-tuned hyperparameter, which is then inferred to the target model. The scaled map divides the region between the safe area limit and the low/high limit into subregions, each representing a key, with the corresponding value being the tuned hyperparameter based on a range of $\alpha \in (0,1]$. Higher values of $\alpha$ (e.g., close to $\mathbf{1}$) imply stronger adaptation and are assigned to regions nearer to the outer low limit, whereas moderate values (e.g., around $\mathbf{0.5}$) indicate minor deviations within the incremental zone, corresponding to points closer to the safe area boundary. This tradeoff prevents overfitting while ensuring adaptability.

\begin{figure}[ht]
    \centering
    \begin{minipage}{\columnwidth}
        \centering
        \includegraphics[width=\columnwidth,keepaspectratio]{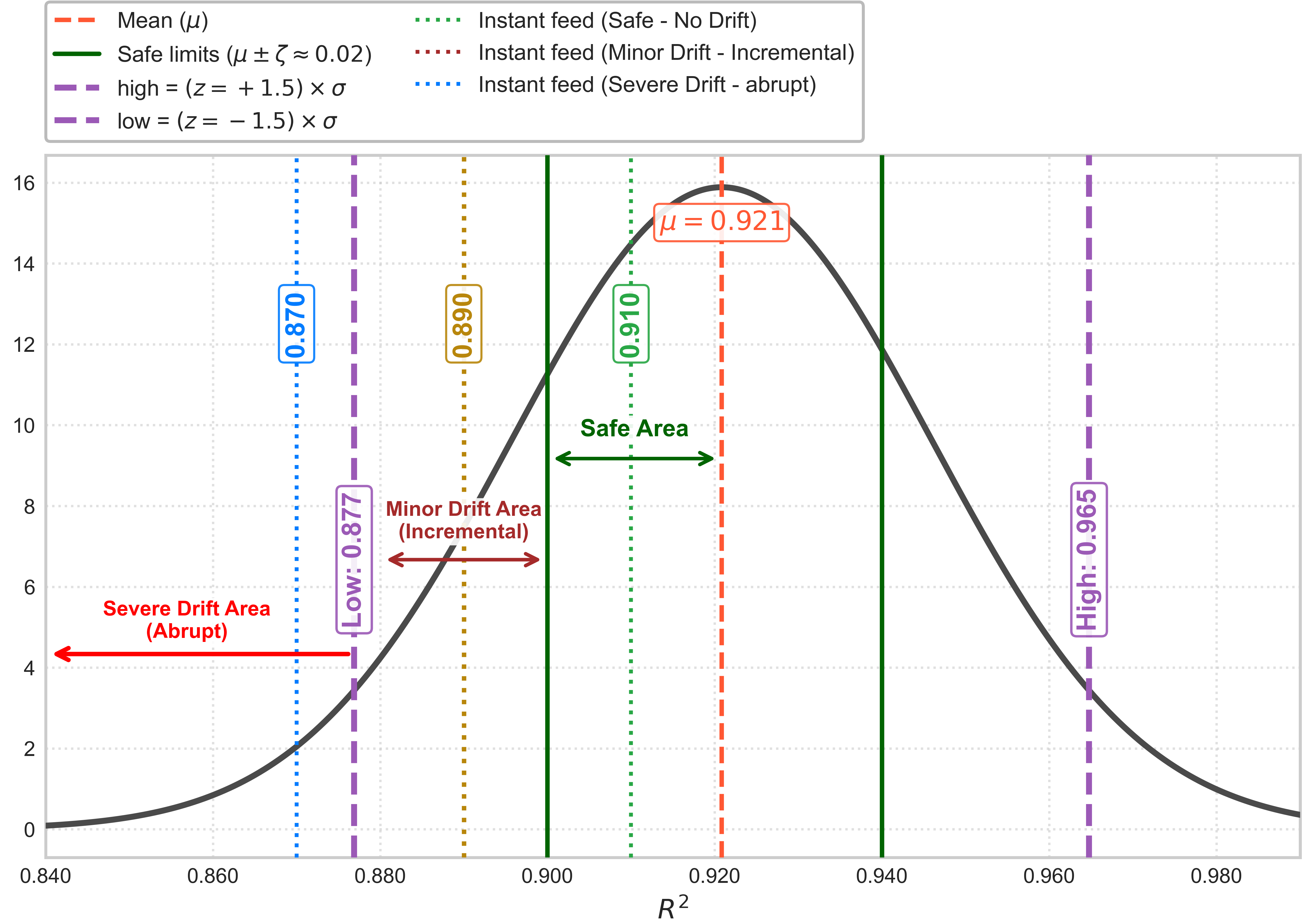}
        \caption{\centering Limits – Gauss Dist. of KPI-Win on KPI = Accuracy.}
        \label{fig:003_SCCM_LIMITS}
    \end{minipage}
\end{figure}

\begin{figure}[ht]
  \centering
  \includegraphics[width=\linewidth,keepaspectratio]{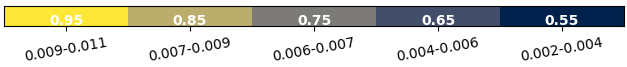}
  \caption{Scale Map}
  \label{fig:004_SCCM_SCALE}
\end{figure}

The mechanism adapts hyperparameters according to the observed drift magnitude, defined as 
$\text{DM} = \mu_{\scriptscriptstyle\text{KPI}} - \text{CUR-MB}_{\scriptscriptstyle\text{KPI}}$, 
where $\mu_{\scriptscriptstyle\text{KPI}}$ is the mean of the KPI window and 
$\text{CUR-MB}_{\scriptscriptstyle\text{KPI}}$ is the KPI value from the current mini-batch. 
To differentiate between stable performance, incremental drift, and abrupt drift, three boundaries are established. 
The inner safe band, $[\mu_{\scriptscriptstyle\text{KPI}} - \zeta,\, \mu_{\scriptscriptstyle\text{KPI}} + \zeta]$, 
where $\zeta$ is a user-defined parameter (set to 0.005 in our experiments when KPI=ACC is utilized), absorbs minor fluctuations that do not warrant adaptation. 
Outside this band, the outer limits are determined by $\tau = z \times \sigma$, where $\sigma$ is the standard deviation of the KPI window 
and $z$ is derived from a user-specified false-alarm probability $\rho$ (e.g., $z=1.5$). 
These yield thresholds $\text{low} = \mu_{\scriptscriptstyle\text{KPI}} - \tau$ and 
$\text{high} = \mu_{\scriptscriptstyle\text{KPI}} + \tau$. 
Observations within the safe band are treated as stable, those between the safe band and the outer limits are classified as incremental drift, 
and those beyond the outer limits as abrupt drift. 
Figure~\ref{fig:003_SCCM_LIMITS} depicts the boundary construction, while Figure~\ref{fig:004_SCCM_SCALE} illustrates how the dynamically constructed 
scale map leverages this framework to select the appropriate $\alpha$ for guiding model adaptation.

\begin{algorithm}
\caption{\modelname}
\label{alg:olcwaalg}
\footnotesize
\begin{algorithmic}[1]

\State \textbf{Input:} Data stream $\{\mathbf{X}, \mathbf{y}\}$, $\alpha \in (0,1]$, $\rho \in (0,1]$, $\textit{kpi} \subseteq \{\text{Acc, Loss, etc.}\}$

\State $\mathbf{W}_{\text{base}} \gets \arg\min\limits_{\mathbf{w}} \mathcal{L}_{\text{NLL}}(\mathbf{X}_{\text{base}}, \mathbf{y}_{\text{base}})$
\makebox[0pt][l]{\hspace{1.25cm}\Comment{\scriptsize \parbox{7cm}{\textcolor{blue}{$\mathbf{X}_{\text{base}}, \mathbf{y}_{\text{base}}$ represent the first mini-batch}}}}

\State Initialize an empty list: $L \gets [\;]$
\makebox[0pt][l]{\hspace{2cm}\Comment{\scriptsize \textcolor{blue}{$L$ is the KPI window}}}

\For{$t \gets 1$ to $T$}

    \State $\mathbf{W}_{\text{inc}} \gets \arg\min\limits_{\mathbf{w}} \mathcal{L}_{\text{NLL}}(\mathbf{X}_{\text{inc}}, \mathbf{y}_{\text{inc}})$

    \State $\mathbf{P}_{\text{int}} \gets \method{FindIntersectPoint}(\mathbf{W}_{\text{base}}, \mathbf{W}_{\text{inc}})$

    \State $\mathbf{V}_{\text{base}} \gets \method{ComputeNormVector}(\mathbf{W}_{\text{base}})$
    \State $\mathbf{V}_{\text{inc}} \gets \method{ComputeNormVector}(\mathbf{W}_{\text{inc}})$

    \State $\mathbf{V}_{\text{base}} \gets \frac{\mathbf{V}_{\text{base}}}{\|\mathbf{V}_{\text{base}}\|_2}$ \quad
           $\mathbf{V}_{\text{inc}} \gets \frac{\mathbf{V}_{\text{inc}}}{\|\mathbf{V}_{\text{inc}}\|_2}$

    \State $\mathbf{V}_{\text{avg}} \gets (1 - \alpha)\,\mathbf{V}_{\text{base}} + \alpha\,\mathbf{V}_{\text{inc}}$

    \State $\mathcal{H}_{\text{base}} \gets \method{DefineHyperplane}(\mathbf{V}_{\text{avg}}, \mathbf{P}_{\text{int}})$
    \makebox[0pt][l]{\hspace{.7cm}\Comment{\scriptsize \textcolor{blue}{Decision boundary defined}}}

    \State $\mathbf{W}_{\text{base}} \gets \begin{bmatrix}\mathbf{V}_{\text{avg}} \\ -\mathbf{V}_{\text{avg}}^{\!\top}\mathbf{P}_{\text{int}}\end{bmatrix}$\makebox[0pt][l]{\hspace{3.2cm}\Comment{\scriptsize \textcolor{blue}{Equivalent parameter vector}}}

    \State $\mathbf{MB}_{\text{kpi}} \gets \method{MbKPI}(\mathbf{X}_{\text{inc}}, \mathbf{y}_{\text{inc}}, \mathbf{W}_{\text{base}}, \textit{kpi})$

    \State \method{AddItem}($L$, $\mathbf{MB}_{\text{kpi}}$)

    \State \scriptsize \texttt{\textcolor{blue}{/* Start hyperparameter optimization once the}}
    \\\hspace{22pt}\texttt{\textcolor{blue}{KPI window reaches its threshold size. */}}

    \State $ \tau, \mu, \sigma, \text{low}, \text{high}, \text{DM} \gets \method{MeasureKPIs}(L, z)$ 
    \Comment{\parbox[t]{5.5cm}{\textcolor{blue}{with $z \gets \Phi^{-1}(1 - \rho)$, \\the threshold corresponding to \\false-alarm probability $\rho$}}}

    \State $\textit{drift} \gets \method{DetectDrift}(\mu, L(-1), \tau)$ 

    \If{\textit{drift}}
        \State \method{RemoveItem}($L$, $-1$)
        \State $\text{SM} \gets \method{DefineScaleMap}(\mu, \text{low}, \text{high})$
        \makebox[0pt][l]{\hspace{1.17cm}\Comment{\scriptsize \textcolor{blue}{SM is the scaled map}}}
        
        \State $\alpha' \gets \method{TuneHyperparams}(\text{SM}, \text{DM})$

        \State $\mathbf{V}_{\text{avg}} \gets (1 - \alpha')\,\mathbf{V}_{\text{base}} + \alpha'\,\mathbf{V}_{\text{inc}}$
        \State $\mathcal{H}_{\text{base}} \gets \method{UpdateHyperplane}(\mathbf{V}_{\text{avg}}, \mathbf{P}_{\text{int}})$
        \makebox[0pt][l]{\hspace{.5cm}\Comment{\scriptsize \textcolor{blue}{\textcolor{blue}{Updated decision boundary}}}}
        
        \State $\mathbf{W}_{\text{base}} \gets \begin{bmatrix}\mathbf{V}_{\text{avg}} \\ -\mathbf{V}_{\text{avg}}^{\!\top}\mathbf{P}_{\text{int}}\end{bmatrix}$ \makebox[0pt][l]{\hspace{3.07cm}\Comment{\scriptsize \textcolor{blue}{\textcolor{blue}{Equivalent parameter vector}}}}
    \EndIf    

\EndFor
\State \textbf{return} $\mathbf{W}_{\text{base}}$
\end{algorithmic}
\end{algorithm}

However, using a fixed smoothing factor \(\alpha\) presents two critical challenges. First, a high smoothing factor can result in \textit{overfitting}, where the model becomes excessively sensitive to recent changes, potentially capturing noise instead of meaningful trends. Furthermore, it increases the risk of \textit{catastrophic forgetting}, where the model fails to retain valuable past knowledge, leading to suboptimal long-term performance. Second, a low smoothing factor will result in very low adaptability in responding to rapid changes in the data stream, as it places greater emphasis on historical data while minimizing the influence of recent observations. This can lead to delayed adjustments in dynamic environments, reducing the model's ability to effectively capture and react to sudden shifts or concept drifts. In response, the proposed \modelname balances \textit{adaptiveness} (responding to new changes) with \textit{robustness} (preserving learned knowledge), ensuring responsiveness to evolving data while maintaining stability.

\begin{figure}[ht]
  \centering
  \includegraphics[width=0.65\textwidth, height=0.2\textheight]{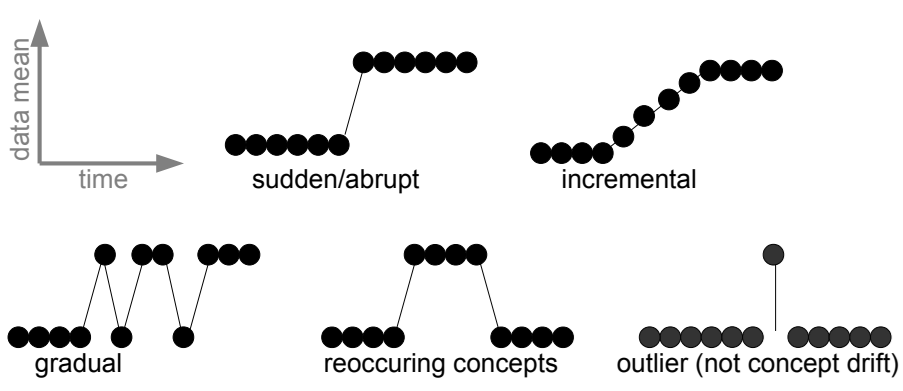}
  \caption{Concept Drift Types \cite{gama2014survey}}
  \label{fig:001_drift_types}
\end{figure}

Concept drift can be broadly categorized into four primary types: \textit{sudden}, \textit{gradual}, \textit{incremental}, and \textit{recurring}, as discussed by Webb et al.~\cite{webb2016characterizing} and Gama et al.~\cite{gama2014survey}, and illustrated in Figure~\ref{fig:001_drift_types}. 
While these categories describe distinct temporal patterns of change, such distinctions generally require a retrospective, \textit{post hoc} analysis of the complete data stream to identify long-term distributional trends.
In an online learning context, however, the learner has access only to recent observations at time $t$, so its goal shifts from retrospective characterization to immediate reaction. 
From this real-time perspective, the subtle boundaries among drift types can be consolidated into two operational categories based on instantaneous severity: 
\textbf{abrupt (severe)} and \textbf{incremental (mild)} drift. 
This simplification is theoretically justified because a gradual drift manifests as a sequence of small abrupt shifts, while a recurring drift appears as an abrupt reversion to a previously observed concept. 

Accordingly, the \modelname\ framework continuously monitors a bounded KPI window (KPI-Win) to track the statistical stability of a chosen performance metric. 
Instantaneous KPI values that remain within the \textit{safe band} $[\mu-\zeta,\ \mu+\zeta]$ are considered stable. 
Deviations \textbf{beyond} this band indicate drift: points lying between the safe band and the outer threshold ($\pm\tau=z\sigma$) are classified as \textbf{incremental drift}, whereas points exceeding $\pm\tau$ are treated as \textbf{abrupt drift}. 
Upon detection, the model automatically re-optimizes its smoothing factor $\alpha$, increasing $\alpha$ under abrupt drift to enable faster adaptation and decreasing it under incremental drift to preserve stability. 
This mechanism enables \textbf{adaptive, tuning-free drift handling in real time}, seamlessly integrated with the memory-based hyperparameter tuning described above.

As shown in Algorithm \ref{alg:olcwaalg}, \modelname utilizes an initial smoothing factor \( \alpha \), which influences the performance of the initial base model. Additionally, for each incoming batch at time \( t \), the initial weight vector is computed using this smoothing factor. If drift is detected, the model undergoes an optimization process, updating \( \alpha \) to \( \alpha' \) to adapt accordingly. 

Notably, the Memory-Based Hyperparameter Tuning Mechanism is a pluggable component that can be disabled. This is particularly useful in confidence-based scenarios where older data points are deemed more reliable and investigate the convergence speed of the models. In this setting,  the user sets a low initial value for \( \alpha \), allowing the model to retain this fixed value across all iterations without dynamic tuning.

\vspace{0.5\baselineskip}
\noindent\textit{Mitigating False Positives:} 
False positives are detections that signal drift when the data stream is still in control relative to the current baseline window, meaning the KPI distribution has not truly changed. In online learning this matters because each false alarm can trigger needless adaptation, retuning, or model rewrites, which destabilizes training, burns compute, and can degrade accuracy by chasing noise instead of signal. At a fine level, an abrupt false positive occurs when a benign fluctuation crosses the one-sided low/high limit on the monitored tail, while an incremental false positive occurs when a benign point lands between the safe boundary and that one-sided low/high limit; both lead the system to act even though the underlying concept is unchanged. Repeated false positives cause parameter oscillations, loss of calibration for hyperparameters such as the adaptation weight, potential forgetting of useful prior structure, and alert fatigue for operators. Mitigating false positives therefore requires explicit control of tail risk on the monitored side of the KPI, careful use of a safe band to absorb small deviations, and clear rules that separate benign variability from actionable drift. The objective is to set the one-sided low/high limits using a threshold derived from $\rho$, such that the false-positive rate equals the user-specified probability $\rho$ (e.g. $\rho$ = .01). These limits are recalibrated as $\mu$ and $\sigma$ update over time, while $\rho$ itself remains fixed.

We mitigate false positives by calibrating the limits (low/high) to a \textit{target constant false-alarm rate} $\rho$. After updating the KPI window $L$, we compute the window mean $\mu$ and standard deviation $\sigma$, retain the safe band $[\mu-\zeta,\ \mu+\zeta]$, and set the low/high limits via $z=\Phi^{-1}(1-\rho)$ and $\tau=z\,\sigma$. For KPIs where higher is better we monitor the lower tail and use $\text{low}=\mu-\tau$; for KPIs where lower is better we monitor the upper tail and use $\text{high}=\mu+\tau$. $\Phi$ denotes the cumulative distribution function (CDF) of the standard normal variable $Z\sim\mathcal{N}(0,1)$. Specifically, $\Phi(z)$ gives the probability that a standard normal observation falls at or below a threshold $z$, i.e., the proportion of the bell curve to the left of $z$. Its inverse, $\Phi^{-1}(q)$, performs the reverse mapping: it returns the quantile $z$ such that exactly a fraction $q$ of the distribution lies to its left. Accordingly, setting $z=\Phi^{-1}(1-\rho)$ selects the cutoff point on the standard normal curve such that only a fraction $\rho$ of the distribution remains in the monitored tail. In other words, the threshold is chosen so that the probability of a false alarm under in-control conditions equals the target false-alarm rate $\rho$.

Points between the safe boundary and the one-sided outer limit indicate incremental drift, and points beyond the one-sided outer limit indicate abrupt drift. Under in-control conditions the probability that a benign point crosses the abrupt limit equals $\rho$ by construction, while $\zeta$ serves as a tunable buffer that controls the trade-off between sensitivity and stability by governing how often incremental responses occur. When Gaussianity is doubtful, $\Phi^{-1}(1-\rho)$ can be replaced by the empirical one-sided $\rho$-quantile of recent residuals without changing the procedure \cite{scharf1991statistical, palama2016multistatic, umsonst2022finite}.

To illustrate how this construction mitigates false positives, recall that choosing $z=\Phi^{-1}(1-\rho)$ fixes a cutoff that leaves exactly a fraction $\rho$ of the standard-normal distribution in the monitored tail. Scaling this cutoff by the standard deviation $\sigma$ then maps the threshold into KPI units. The outcome is a one-sided \textit{constant false-alarm rate (CFAR) threshold} \cite{rohling2007radar}, which guarantees that, under in-control conditions, the probability of a benign observation crossing the abrupt limit equals $\rho$ by design. In parallel, the safe band $\zeta$ introduces a dead zone that absorbs small fluctuations, reducing nuisance incremental drift flags and ensuring that only meaningful deviations trigger adaptation.

In an illustrative example, consider the Accuracy KPI with parameters $\mu = 0.921$, $\sigma = 0.010$, $\zeta = 0.020$, and $\rho = 0.01$. The calibration yields $z = \Phi^{-1}(0.99) \approx 2.326$ and $\tau = 0.023$, giving a safe band of $[0.900, 0.940]$. Since higher values are favorable for accuracy, the one-sided monitoring limit is $\text{low} = 0.877$. A reading of $0.91$ lies within the safe band and is therefore stable, while a reading of $0.890$ falls between the safe band and the low limit and is classified as incremental drift. A reading of $0.870$ falls below the low limit and is classified as abrupt drift. In contrast, a reading of $0.950$ lies above the safe band on the non-monitored side and is interpreted as an improvement rather than a trigger. Under stable conditions, the probability that a benign point crosses the abrupt limit equals $\rho = 0.01$, as illustrated in Figure \ref{fig:003_SCCM_LIMITS}.

\subsection{Theoretical Analysis of Convergence}
This analysis provides a formal examination of the convergence properties of the proposed $\text{\modelname}$ framework, demonstrating its stability and asymptotic behavior under continuous online updates.
\subsubsection{Model Description}
The \modelname model incrementally learns from streaming data by maintaining two decision boundaries: the base decision boundary, \( f_{\text{base}}(x) \), which encapsulates the accumulated knowledge from past observations, and the incremental decision boundary, \( f_{\text{inc}}(x) \), which reflects the knowledge derived from the current mini-batch (instant feed) at time \( t \), where \( x \) denotes the feature vector. The corresponding norm vectors of these boundaries, denoted as \( V_{\text{base}} \) and \( V_{\text{inc}} \), respectively, are integrated using the EWMA technique. The dynamic weighting factor \( \alpha \in (0,1] \) serves as a smoothing parameter, assigning a weight of \( \alpha \) to the incremental norm vector, emphasizing the contribution of the most recent data, while the remaining weight \( (1 - \alpha) \) accounts for the historical knowledge retained in the base model. The combined norm vector \( V_{\text{Avg}} \) is iteratively updated, leveraging the intersection point \( P_{\text{int}} \) to redefine the base norm vector, with the objective of progressively aligning with \( V_{\text{inc}} \) as the number of observations increases. \textit{Assumptions:}

\begin{enumerate}[labelsep=0.8em]
    \item \textit{Linearity:} The relationship between the features and the target variable follows a linear model.    
    \item \textit{Concept Stability} \cite{buzmakov2014concept}\textbf{:} While the overall data distribution may drift over time, each mini-batch is assumed to represent a stable concept, enabling effective adaptation without ambiguity.
    \item \textit{Boundedness of Data:} Input features and target variable are confined within a finite range, ensuring stability and preventing unrealistic predictions beyond the defined input domain.
    \item \textit{Continuity:} Decision boundaries are continuous functions.
\end{enumerate}

\subsubsection{Convergence Criterion}
Our objective is to demonstrate the asymptotic convergence of the \modelname model, specifically focusing on the intersection point and the weighted average norm vector, as the number of observations approaches infinity.
% \\\\
% \noindent
\begin{enumerate}[labelsep=0.8em]
    \item \textbf{Step 1: Convergence of Incremental Norm Vector \( V_{\text{inc}} \):}
    
    Under the assumption of concept stability \cite{buzmakov2014concept}, the norm vector \( V_{\text{inc}} \) of the new incoming mini-batches (instant feeds) is expected to adapt and approximate optimal values over time as \( t \rightarrow \infty \), facilitating improved classification performance for any given instant feed at time \( t \).

    \item \textbf{Step 2: EWMA-Based Integration and Computation of the Weighted Average Norm Vector:}

    Upon receiving the classification decision boundaries (hyperplanes) from both the base and incremental models at every time point \( t \), where the base model encapsulates the past accumulated knowledge from all previous observations and the incremental model reflects the knowledge derived from the current mini-batch, the \modelname framework computes their corresponding norm vectors, denoted as \( V_{\text{base}} \) and \( V_{\text{inc}} \). To effectively capture the evolving decision boundaries, \modelname combines these norm vectors to derive a weighted average norm vector, \( V_{\text{Avg}} \), at each iteration \( t \). This vector encapsulates both the direction and magnitude of the evolving classification model.
    
    The integration process employs the Exponentially Weighted Moving Average (EWMA) approach, where the weighting factor \( \alpha \) determines the relative importance of the most recent incremental mini-batches (instant feeds), while \( 1 - \alpha \) accounts for previously accumulated knowledge. The weighting factor \( \alpha \) serves as an adaptation parameter, allowing the model to balance the responsiveness to new information with the retention of previous learning. The weighted average norm vector, \( V_{\text{Avg}} \), is computed as follows:
    \[
    V_{\text{Avg}} = \alpha \cdot V_{\text{inc}} + (1 - \alpha) \cdot V_{\text{base}}
    \]

    \item \textbf{Step 3: Convergence of \modelname Model:}
    
    \textbf{Convergence of} \( V_{\text{Avg}} \) \textbf{to} \( V_{\text{inc}} \): 
    Let \( \Delta_t = \| V_{\text{Avg}} - V_{\text{inc}} \| \) represent the norm of the difference between the weighted average norm vector and the incremental norm vector associated with the evolving decision boundaries. We aim to prove that \( \Delta_t \rightarrow 0^+ \) as \( t \rightarrow \infty \), as formalized in Theorem~\ref{th:ewma_convergence}.

\end{enumerate}

\begin{theorem}
Let \( t \) denote the number of iterations. Let \( \alpha \) denote the smoothing parameter. As \( t \rightarrow \infty \), the EWMA converges to the true mean or central tendency of the observed process.
\label{th:ewma_convergence}
\end{theorem}

\subsubsection{Proof Sketch}
To establish the convergence behavior of the proposed \modelname framework, we outline the key analytical steps leading to the convergence of the weighted average norm vector \( V_{\text{Avg}} \) toward the incremental norm vector \( V_{\text{inc}} \). The proof proceeds through the following steps:

\begin{enumerate}[labelsep=0.8em]
    \item In \modelname, the weighted average norm vector \( V_{\text{Avg}} \) is obtained by combining the norm vectors of base and incremental regression hyperplanes using EWMA with dynamic weighting factor \( \alpha \) selected adaptively from a predefined range (0,1].
    \item Therefore, \( V_{\text{Avg}} = \alpha \cdot V_{\text{inc}} + (1 - \alpha) \cdot V_{\text{base}} \).
    \item The weighted average norm vector is computed iteratively using the EWMA formula to ensure an adaptive balance between recent and historical information.
\end{enumerate}
Systematically incorporating all past observations, an iterative  EWMA formulation results in the following expansion:
\[
\bar{X}_t = (1 - \alpha_t) \bar{X}_{t-1} + \alpha_t X_t
\]
\[
\bar{X}_t = (1 - \alpha_t) \left[(1 - \alpha_{t-1}) \bar{X}_{t-2} + \alpha_{t-1} X_{t-1} \right] + \alpha_t X_t
\]
\[
\bar{X}_t = (1 - \alpha_t)(1 - \alpha_{t-1}) \bar{X}_{t-2} + \alpha_{t-1} (1 - \alpha_t) X_{t-1} + \alpha_t X_t
\]
\vspace{-20pt}
\begin{multline*}
    \bar{X}_t = (1 - \alpha_t)(1 - \alpha_{t-1}) \left[(1 - \alpha_{t-2}) \bar{X}_{t-3} + \alpha_{t-2} X_{t-2} \right] + \\ \alpha_{t-1} (1 - \alpha_t) X_{t-1} + \alpha_t X_t
\end{multline*}
\vspace{-20pt}
\begin{multline*}
    \bar{X}_t = (1 - \alpha_t)(1 - \alpha_{t-1})(1 - \alpha_{t-2}) \bar{X}_{t-3} + \alpha_{t-2} (1 - \alpha_t)\\(1 - \alpha_{t-1}) X_{t-2} + \alpha_{t-1} (1 - \alpha_t) X_{t-1} + \alpha_t X_t
\end{multline*}
\[
\vdots
\]
\vspace{-20pt}
\begin{multline*}
\bar{X}_t = \prod_{j=1}^{t} (1 - \alpha_j) \bar{X}_{0} + \sum_{k=0}^{t-1} \left( \alpha_{t-k} \prod_{j=t-k+1}^{t} (1 - \alpha_j) \right) X_{t-k}.    
\end{multline*}
Through adaptive exponential smoothing, past observations fade at rates controlled by the dynamic \(\alpha_t\). Higher \(\alpha_t\) prioritizes recent data for faster adaptation, while lower \(\alpha_t\) retains historical influence longer, balancing responsiveness and stability. Applying the same principles to the \modelname model:
\begin{multline*}
    V_{\text{Avg}, t} = \underbrace{\prod_{j=1}^{t} (1 - \alpha_j) V_{\text{base}}}_{\text{First term}}
    +\underbrace{\sum_{k=0}^{t-1} \left( \alpha_{t-k} \prod_{j=t-k+1}^{t} (1 - \alpha_j) \right) V_{\text{inc}, t-k}}_{\text{Second term}}.
\end{multline*}
\\\\
\noindent
\textbf{First Term Analysis:}

The first term in the equation represents the contribution of \(V_{\text{base}}\), while the second term captures the accumulated effect of the incremental norm vectors \(V_{\text{inc}, t-k}\). The factor \((1 - \alpha_j)\) denotes the fraction of past information retained at each step, and its product \(\prod_{j=1}^{t} (1 - \alpha_j)\) reflects the cumulative retention over multiple time steps. If \(\alpha_j\) were constant (\(\alpha_j = \alpha\)), this product would follow \((1 - \alpha)^t\), exhibiting an exponential decay toward zero for \(0 < \alpha \leq 1\). However, since \(\alpha_j\) is dynamically selected, a more general bound is required to account for its adaptive nature.
\\\\
\noindent
\textit{Lower Bound of the First Term:}  
Since \(\alpha_j\) is selected from \((0,1]\), it follows that \(0 < (1 - \alpha_j) \leq 1\). Consequently, their product remains non-negative and at most 1:  
\begin{equation*}
0 < \prod_{j=1}^{t} (1 - \alpha_j) \leq 1.    
\end{equation*}
The lower bound of 0 ensures that \(\prod_{j=1}^{t} (1 - \alpha_j)\) is always nonnegative.
\\\\
\noindent
\textit{Upper Bound of the First Term:}
The upper bound determines the decay rate of \(\prod_{j=1}^{t} (1 - \alpha_j)\) over time. Using the logarithmic inequality \( \ln(1 - x) \leq -x \) for \( 0 \leq x \leq 1 \), we obtain:
\[
\ln \left( \prod_{j=1}^{t} (1 - \alpha_j) \right) = \sum_{j=1}^{t} \ln(1 - \alpha_j) \leq -\sum_{j=1}^{t} \alpha_j.
\]
Exponentiating both sides:
\[
\prod_{j=1}^{t} (1 - \alpha_j) \leq e^{-\sum_{j=1}^{t} \alpha_j}.
\]
This bound ensures \textit{exponential decay}, meaning if \( \sum_{j=1}^{t} \alpha_j \to \infty \) as \( t \to \infty \), then \( \prod_{j=1}^{t} (1 - \alpha_j) \to 0 \), causing \( V_{\text{base}} \) to vanish:
\[
   \lim_{t \to \infty} \prod_{j=1}^{t} (1 - \alpha_j) V_{\text{base}} = 0.
\]
Thus, the model fully transitions to adaptive weighting based on incremental norm vectors.
\\\\
\noindent
\textbf{Second Term Analysis:}

The second term represents the accumulated influence of the past incremental norm vectors, where each contribution is weighted by $\alpha_{t-k}$ and discounted by the decay factor $\prod_{j=t-k+1}^{t} (1 - \alpha_j)$. We establish the following key result on the convergence of the \textbf{sum of weights}:
\[
\sum_{k=0}^{t-1} \left( \alpha_{t-k} \prod_{j=t-k+1}^{t} (1 - \alpha_j) \right) = 1 - \prod_{j=1}^{t} (1 - \alpha_j).
\]
Taking the limit as $ t \to \infty $, and noting that $\prod_{j=1}^{t} (1 - \alpha_j) \to 0$ (as established in the First Term Analysis), we obtain:
\[
\lim_{t \to \infty} \sum_{k=0}^{t-1} \left( \alpha_{t-k} \prod_{j=t-k+1}^{t} (1 - \alpha_j) \right) = 1.
\]
Thus, the weighted sum of past incremental contributions \textbf{fully determines the model} (since the first term vanished), leading to the convergence of the norm vector:
\[
\lim_{t \to \infty} V_{\text{Avg}, t} = \sum_{k=0}^{\infty} \left( \alpha_{t-k} \prod_{j=t-k+1}^{t} (1 - \alpha_j) \right) V_{\text{inc}, t-k}.
\]
Since $\alpha_{t-k}$ controls the weight assigned to each past update while the decay factor reduces the influence of older updates, and given that we have established the sum of weights converges to one, the vector $V_{\text{Avg}, t}$ is guaranteed to converge to the central tendency of the sequence of $V_{\text{inc}}$ vectors.
\\
\\
\noindent
\textbf{Final Convergence Results:} 

This result confirms that $V_{\text{Avg}, t}$ is now entirely determined by the continuous stream of incremental updates, meaning the model has fully transitioned to relying on new data (as the influence of the initial $V_{\text{base}}$ vanishes).

The $\sum \text{weights} = 1$ property guarantees the following convergence under stable conditions:

\[
\Delta_t = \| V_{\text{Avg}, t} - V_{\text{inc}, t} \|.
\]

If concept drift ceases and the optimal incremental norm vector converges to a fixed value $V_{\text{opt}}$ (i.e., $\lim_{t \to \infty} V_{\text{inc}, t} = V_{\text{opt}}$), then the total weight summing to one ensures that $V_{\text{Avg}, t}$ will also converge to that optimal norm vector:
\[
\lim_{t \to \infty} V_{\text{Avg}, t} = V_{\text{opt}}.
\]
Consequently, under this stable condition, the difference vanishes:
\[
\lim_{t \to \infty} \Delta_t = 0.
\]
This guarantees that the decision boundary smoothly transitions, incorporating past knowledge while ensuring asymptotic stability and responsiveness whenever the underlying concept stabilizes.

%@@@@@@@@@@@@@@@@@@@@@@@@@@@@@@@@@@@@@@@
\subsection{Theoretical Foundations and Regret Bound Analysis}
\label{sec:regret_analysis}

\subsubsection{Non-Asymptotic Performance Guarantees and Assumptions}
To provide a rigorous, non-asymptotic characterization of the finite-sample behavior of the $\text{\modelname}$ framework, we analyze its cumulative performance using the \textbf{static regret} metric, grounded in Online Convex Optimization (OCO) theory.

The cumulative regret $R_T$ over $T$ rounds compares the total loss incurred by the online learner with that of the best fixed model in hindsight. Let $w^*$ denote the optimal static comparator that minimizes the cumulative loss:
\[
w^* = \arg\min_{w \in \mathcal{W}} \sum_{t=1}^{T} L_t(w),
\]
where $L_t(w)$ is the convex instantaneous loss at round $t$. The cumulative regret is defined as
\[
R_T = \sum_{t=1}^{T} L_t(w_t) - \sum_{t=1}^{T} L_t(w^*).
\]
Our objective is to show that $\text{\modelname}$ achieves a \textbf{sublinear regret bound}, $R_T = O(\sqrt{T})$, which guarantees that the average regret $R_T/T$ converges to zero as $T \to \infty$. This ensures that, over time, the performance of $\text{\modelname}$ asymptotically matches that of the best fixed model.

\paragraph{Equivalence of Parameter Representations.}
The geometric representation in $\text{\modelname}$—comprising the norm vector $V$ and intersection point $P$—is analytically equivalent to the standard OCO weight–bias representation via the mapping $w \equiv V$ and $b \equiv -V^{\top}P$. Because the exponentially weighted average (EWMA) of the norm vectors, $V_{\text{avg}} = (1 - \alpha_t)V_{\text{base}} + \alpha_t V_{\text{inc}}$, directly corresponds to the updated parameter vector $w_{t+1}$, the subsequent analysis adopts the standard OCO notation $w$ for consistency.

\paragraph{Core Assumptions.}
The following mild and standard OCO assumptions are invoked to ensure a valid regret guarantee:
\begin{enumerate}[labelsep=0.8em]
    \item \textit{Convex Loss:} Each instantaneous loss $L_t(w)$ is convex in $w \in \mathcal{W}$.
    \item \textit{Bounded Domain ($D$):} The feasible parameter space $\mathcal{W}$ is a convex set with finite diameter $D$, satisfying $\|u - v\| \le D$ for all $u, v \in \mathcal{W}$.
    \item \textit{Bounded Gradients ($G$):} The gradients of the loss are bounded, i.e., $\|\nabla L_t(w)\| \le G$ for all $w \in \mathcal{W}$ and all rounds $t$.
    \item \textit{Bounded Adaptive Weight:} The adaptive coefficient $\alpha_t$ that balances stability and adaptability satisfies $0 < \alpha_{\min} \le \alpha_t \le \alpha_{\max} \le 1$.
\end{enumerate}

\subsubsection{Adaptive Gradient-Based Interpretation of the Update Rule}
The update mechanism of $\text{\modelname}$ can be interpreted as an adaptive gradient-based procedure. At each round $t$, the parameter vector is updated as
\[
w_{t+1} = (1 - \alpha_t)w_t + \alpha_t \tilde{w}_t,
\]
where $\tilde{w}_t$ is the parameter estimated from the current mini-batch. Rewriting the update as $w_{t+1} = w_t - \alpha_t (w_t - \tilde{w}_t)$ highlights that the displacement $(w_t - \tilde{w}_t)$ represents the adjustment direction derived from the most recent data. Since $\tilde{w}_t$ minimizes $L_t(w)$, this direction aligns with the negative gradient $-\nabla L_t(w_t)$, leading to the equivalent form
\[
w_{t+1} = w_t - \alpha_t \nabla L_t(w_t),
\]
where $\alpha_t$ acts as the \emph{adaptive step size} controlling the magnitude of change. This formulation connects $\text{\modelname}$ to the adaptive-gradient family of online convex optimization methods.

\subsubsection{Regret Bound Theorem and Proof Sketch}
The regret bound is derived by examining the geometric progress of the parameter updates and applying standard convexity arguments.

\paragraph{The One-Step Progress Inequality.}
Starting from the adaptive update rule and expanding the squared distance to the optimal comparator $w^*$ yields
\[
\|w_{t+1} - w^*\|^2 = \|w_t - w^*\|^2 - 2\alpha_t \langle \nabla L_t(w_t), w_t - w^* \rangle + \alpha_t^2 \|\nabla L_t(w_t)\|^2.
\]
By convexity, $L_t(w_t) - L_t(w^*) \le \langle \nabla L_t(w_t), w_t - w^* \rangle$. Substituting this into the previous expression provides the fundamental per-round progress relation:
\[
2\alpha_t [L_t(w_t) - L_t(w^*)]
\le
\|w_t - w^*\|^2 - \|w_{t+1} - w^*\|^2 + \alpha_t^2 \|\nabla L_t(w_t)\|^2.
\]
This inequality demonstrates that each update reduces the distance to the optimal solution, up to a small residual term of order $\alpha_t^2$.

\paragraph{Final Regret Bound.}
Summing the above inequalities over $t = 1$ to $T$ telescopes the distance terms, leaving only the initial and final distances. Under Assumptions (A2) and (A3), we obtain
\[
R_T \le \frac{D^2}{2\alpha_{\min}} + \frac{G^2}{2}\sum_{t=1}^{T} \alpha_t^2.
\]
When $\alpha_t$ follows a diminishing schedule proportional to $1/\sqrt{t}$, or remains within a bounded adaptive range $[\alpha_{\min}, \alpha_{\max}]$, the cumulative regret satisfies $R_T = O(\sqrt{T})$. Consequently, the average regret $R_T/T$ vanishes asymptotically, ensuring that $\text{\modelname}$ performs, on average, as well as the optimal static comparator.

\begin{theorem}
\label{th:regret_final}
Under Assumptions (A1)–(A4) and a bounded or diminishing adaptive coefficient $\alpha_t$, the cumulative regret of the $\text{\modelname}$ framework satisfies $R_T = O(\sqrt{T})$. Therefore, the performance of $\text{\modelname}$ asymptotically matches that of the optimal static model while maintaining an adaptive balance between stability and reactivity to non-stationary data.
\end{theorem}

%@@@@@@@@@@@@@@@@@@@@@@@@@@@@@@@@@@@@@@@

\subsection{Complexity Analysis}
\subsubsection{Time Complexity}
\modelname scales \textit{linearly} with respect to feature dimension and mini-batch size, ensuring efficiency in online learning. Its time complexity is bounded by the logit model's \( O(k \cdot i \cdot d) \), where \( k \) is the number of samples per mini-batch, \( d \) the number of features, and \( i \) the number of optimization iterations. The linear system involves only two equations, reducing complexity to constant time \( O(1) \) instead of the usual \( O(n^3) \) for Gaussian elimination \cite{jeannerod2013rank}, where $n$ is the number of equations. EWMA computation and hyperplane updates operate in \( O(d) \). KPI statistics (mean, standard deviation) over a fixed window \( |L| = 31 \) yield constant-time \( O(1) \). Since KPIs are computed per mini-batch, the per-batch complexity is \( O(k \cdot d) \). Drift detection compares KPI means to thresholds in \( O(1) \). If drift is detected, updates include removing the oldest KPI \( O(1) \), scaling map definition \( O(1) \), norm vector update \( O(d) \), and hyperplane update \( O(d) \). Appending a new KPI is also \( O(1) \). Thus, total overhead remains bounded by \( O(k \cdot i \cdot d) \), making \modelname efficient for high-dimensional, fast-streaming data.

\vspace{5pt}
% \stepcounter{subsubsection}
% \noindent\textbf{\thesubsubsection\ Memory Complexity: }  
\subsubsection{Memory Complexity}
The memory usage of \modelname is primarily influenced by two components: the model's weight vector and the KPI-Win buffer used for performance tracking and drift detection. The model maintains a weight vector \(\mathbf{w} \in \mathbb{R}^d\), where \(d\) denotes the number of input features, resulting in a space complexity of \(O(d)\). Additionally, the KPI-Win buffer stores the most recent \(|L|\) entries, each containing \(s\) scalar performance indicators. Typically, \(s = 2\), representing metrics such as accuracy and log loss tracked per mini-batch. This yields a total memory complexity of \(O(d + |L| \cdot s)\), which is compact and scalable, making the method suitable for deployment in high-throughput streaming environments.

\section{Experiments}
This section evaluates performance on synthetic and real-world datasets (Tables \ref{tab:datasets-properties_nc}, \ref{tab:datasets-properties_drift}), covering classification and concept drift with detailed settings and reproducibility guidelines.

\begin{table}[H]
\centering
\setlength{\tabcolsep}{4pt}
\renewcommand{\arraystretch}{1.1}
\caption{Dataset Properties.\\ {\scriptsize S.C: Standard Classification, i.e., Stationary}}
\label{tab:datasets-properties_nc}
\begin{adjustbox}{width=\linewidth} % or \textwidth
\begin{tabular}{lcccccccc}
\toprule
\textbf{Dataset} & \textbf{Type} & \textbf{Points} & \textbf{Dims} & \textbf{Classes} & \textbf{Noise} & \textbf{Train} & \textbf{Test} & \textbf{Focus} \\
\midrule
\textbf{DS1} \cite{OLC_WA_GIT} & Synth. & 1k & 2 & 2 & 10 & 800 & 200 & S.C \\
\textbf{DS2} \cite{OLC_WA_GIT} & Synth. & 10k & 20 & 2 & 20 & 8k & 2k & S.C \\
\textbf{DS3} \cite{OLC_WA_GIT}& Synth. & 10k & 200 & 2 & 25 & 8k & 2k & S.C \\
\textbf{RSIND} \cite{RSIND}& Real & 899 & 7 & 2 & N.A & 720 & 180 & S.C \\
\textbf{CCDD} \cite{CCDD}& Real & 1.5k & 49 & 2 & N.A & 800 & 200 & S.C \\
\textbf{ESPD} \cite{ESPD}& Real & 4.6k & 57 & 2 & N.A & 3.7k & 920 & S.C \\
\textbf{DS7} \cite{OLC_WA_GIT}& Synth. & 1k & 5 & 3 & 10 & 800 & 200 & S.C \\
\textbf{DS8} \cite{OLC_WA_GIT}& Synth. & 10k & 20 & 5 & 20 & 8k & 2k & S.C \\
\textbf{DS9} \cite{OLC_WA_GIT}& Synth. & 10k & 200 & 10 & 25 & 8k & 2k & S.C \\
\textbf{CSCR} \cite{CSCR}& Real & 100k & 21 & 3 & N.A & 80k & 20k & S.C \\
\textbf{HARD} \cite{HARD}& Real & 10.3k & 562 & 6 & N.A & 8.2k & 2.1k & S.C \\
\textbf{HDWD} \cite{HDWD}& Real & 1.8k & 64 & 10 & N.A & 1.4k & 360 & S.C \\
\bottomrule
\end{tabular}
\end{adjustbox}
\end{table}

\subsection{Standard Classification Scenarios}
Standard classification scenarios refer to stationary settings in which the data distribution remains stable, exhibiting no measurable drift over time. In such cases, datasets can be effectively handled by conventional batch classifiers, which are able to achieve high performance.

\textit{Performance Evaluation} 
Table~\ref{tab:algorithm-performance} presents the performance results across datasets listed in Table~\ref{tab:datasets-properties_nc}, revealing several key observations. While PA and OLR perform competitively, their effectiveness hinges on proper tuning of the regularization parameter \(C\) and learning rate \(\lambda\), respectively. Both models struggle on HARD and HDWD. LMS exhibits high sensitivity to \(\lambda\), resulting in unstable and poor outcomes that are hard to correct in online scenarios. PLA consistently underperforms due to its aggressive correction behavior, which leads to overfitting and instability. ONB and VFDT also show inconsistent performance, particularly on CCDD, HDWD, and HARD, often due to dataset-specific challenges. In contrast, \modelname achieves performance on par with the batch model, positioning it as a strong candidate for real-time classification tasks.

\begin{table}[H]
\centering
\setlength{\tabcolsep}{4pt}
\renewcommand{\arraystretch}{1.1}
\caption{\centering Performance Analysis on Normal Classification Scenarios.\\
{\scriptsize Summary of Test Data Accuracy from 5-Fold Cross-Validation}}
\label{tab:algorithm-performance}
\begin{adjustbox}{width=\textwidth} % keep full width like Table 1
\begin{tabular}{lcccccccc}
\toprule
\textbf{Dataset} & \textbf{Batch} & \textbf{PLA} & \textbf{LMS} & \textbf{OLR} & 
\textbf{ONB} & \textbf{PA} & \textbf{VFDT} & \textbf{\modelname} \\
\midrule
\textbf{DS1}   & 0.8784 & 0.8026 & 0.8650 & 0.8728 & 0.8538 & 0.8732 & 0.8732 & \textbf{0.8732} \\
\textbf{DS2}   & 0.8468 & 0.7457 & 0.8448 & 0.8434 & 0.8347 & 0.8388 & 0.8503 & \textbf{0.8452} \\
\textbf{DS3}   & 0.8154 & 0.7108 & 0.7928 & 0.8081 & 0.8042 & 0.8124 & 0.8209 & \textbf{0.8102} \\
\textbf{RSIND} & 0.8651 & 0.7500 & 0.8595 & 0.8658 & 0.8366 & 0.8682 & 0.8366 & \textbf{0.8667} \\
\textbf{CCDD}  & 0.8888 & 0.7914 & 0.8852 & 0.8842 & 0.1335 & 0.8848 & 0.8869 & \textbf{0.8838} \\
\textbf{ESPD}  & 0.9247 & 0.8725 & 0.8809 & 0.9152 & 0.8110 & 0.9198 & 0.7132 & \textbf{0.9215} \\
\textbf{DS7}   & 0.9310 & 0.9132 & 0.8286 & 0.9288 & 0.9282 & 0.9300 & 0.9282 & \textbf{0.9298} \\
\textbf{DS8}   & 0.8378 & 0.6898 & 0.6429 & 0.8373 & 0.8364 & 0.8364 & 0.8364 & \textbf{0.8323} \\
\textbf{DS9}   & 0.7530 & 0.5953 & 0.5010 & 0.7707 & 0.7709 & 0.7699 & 0.7709 & \textbf{0.7590} \\
\textbf{CSCR}  & 0.6594 & 0.4808 & 0.5381 & 0.6430 & 0.6273 & 0.6329 & 0.6815 & \textbf{0.6412} \\
\textbf{HARD}  & 0.9777 & 0.8517 & 0.6603 & 0.8869 & 0.7424 & 0.8811 & 0.7419 & \textbf{0.9757} \\
\textbf{HDWD}  & 0.9659 & 0.8745 & 0.5474 & 0.9087 & 0.0990 & 0.9015 & 0.8930 & \textbf{0.9452} \\
\bottomrule
\end{tabular}
\end{adjustbox}

\vspace{2pt}

\begin{tablenotes}
    \vspace{1.5pt}
    \item \tiny 
    \begin{minipage}{0.90\textwidth} 
    \textbf{*Reproducibility Parameters:}
    
    \textbf{LMS}: ($\lambda$) 
    \parbox[t]{0.85\textwidth}{%
        $\bullet$ [DS1, DS8, DS9] \(\rightarrow 10^{-2}\)  
        $\bullet$ [DS2, DS3, RSIND, CCDD, ESPD, HARD] \(\rightarrow 10^{-3}\) \\ 
        $\bullet$ [DS10, CSCR, HDWD] \(\rightarrow 10^{-4}\)}   
    
    \textbf{OLR}: ($\lambda$) 
    \parbox[t]{0.85\textwidth}{%
        $\bullet$ [DS1, DS8, HARD] \(\rightarrow 10^{-1}\)  
        $\bullet$ [DS2, RSIND, CCDD, ESPD, DS9] \(\rightarrow 10^{-2}\) \\ 
        $\bullet$ [DS3, DS10, CSCR] \(\rightarrow 10^{-3}\)  
        $\bullet$ [HDWD] \(\rightarrow 10^{-4}\)}  
    
    \textbf{PA}: ($C$) 
    \parbox[t]{0.85\textwidth}{%
        $\bullet$ [HARD, HDWD] \(\rightarrow 1\)  
        $\bullet$ [DS1, DS2] \(\rightarrow 10^{-1}\)  
        $\bullet$ [RSIND, ESPD, DS8, DS9] \(\rightarrow 10^{-2}\) \\ 
        $\bullet$ [DS3, CCDD, DS10, CSCR] \(\rightarrow 10^{-3}\)}  

    \textbf{OLC-WA}: 
    \parbox[t]{0.85\textwidth}{%
        $\bullet$ $\alpha = 0.5$ \quad 
        $\bullet$ KPI = ACC \quad 
        $\bullet$ $\rho \in [0.06, 0.006]$}
    \end{minipage}
\end{tablenotes}

\end{table}

\subsection{Concept Drift Scenarios}

\subsubsection{Abrupt Drift}
\textit{Performance Evaluation} 
The models were evaluated on the DS15 to DS18 concept drift datasets (Figure~\ref{fig:exp_abrupt}). \modelname demonstrated the highest resilience to concept drift due to its built-in optimization mechanism. In contrast, N.B. and VFDT performed poorly, exhibiting gradual degradation and requiring substantial portions of the data stream to recover. PA initially struggled but adapted quickly to distributional changes. PLA showed abrupt performance drops, especially during drift events. LMS and OLR encountered early adaptation issues but progressively adjusted to the new distributions across most datasets.

\subsubsection{Incremental Drift}
\begin{figure*}[!htbp]
    \centering
    \begin{subfigure}{0.48\textwidth}
        \centering
        \includegraphics[width=\linewidth, height=4.45cm]{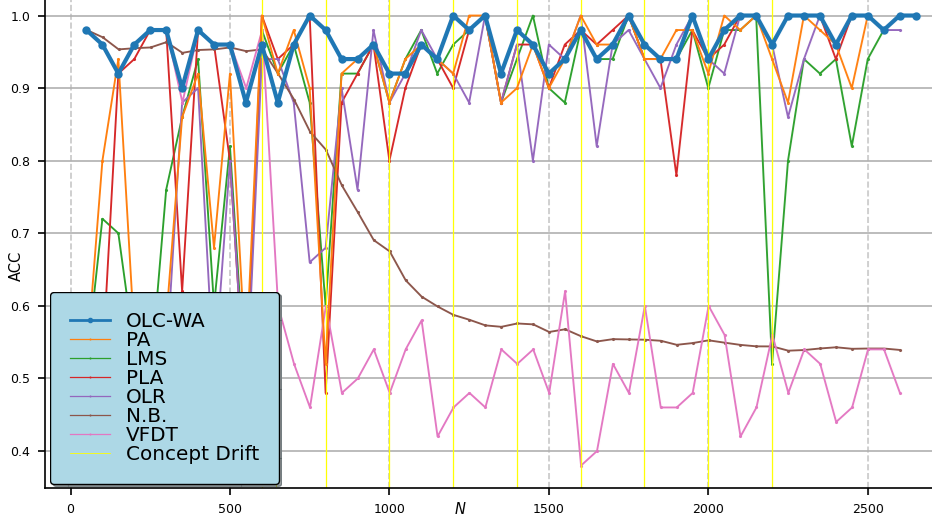}
        \vspace{-.45cm}
        \caption{\scriptsize Models' Performance on DS19}
        \label{fig:sub1}
    \end{subfigure}
    \hfill
    \begin{subfigure}{0.48\textwidth}
        \centering
        \includegraphics[width=\linewidth, height=4.45cm]{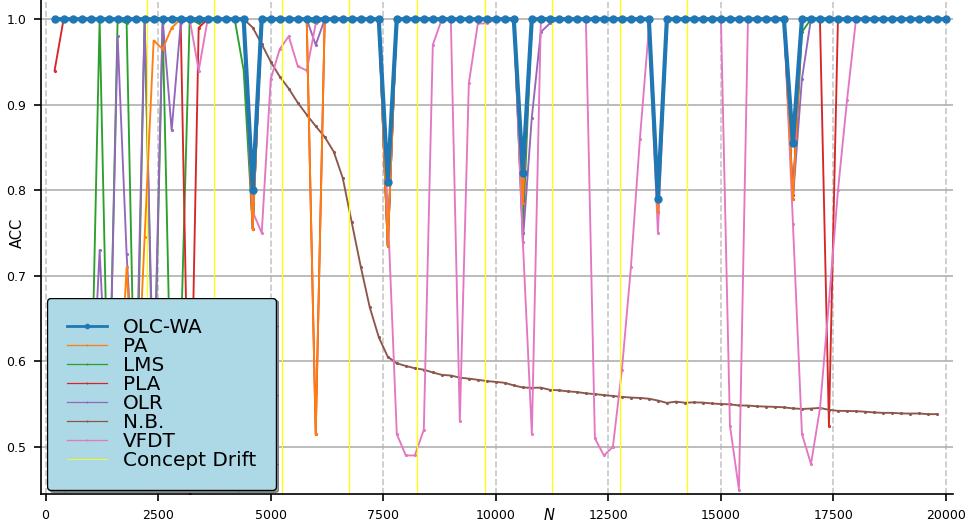}
        \vspace{-.45cm}
        \caption{\scriptsize Models' Performance on DS20}
        \label{fig:sub2}
    \end{subfigure}    
    \begin{subfigure}{0.48\textwidth}
        \centering
        \includegraphics[width=\linewidth, height=4.45cm]{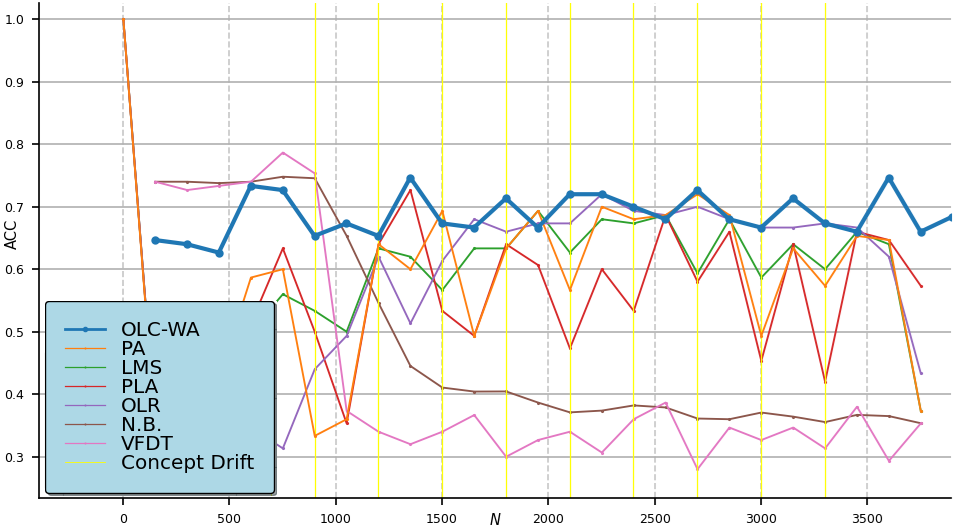}
        \vspace{-.45cm}
        \caption{\scriptsize Models' Performance on DS21}
        \label{fig:sub3}
    \end{subfigure}
    \hfill
    \begin{subfigure}{0.48\textwidth}
        \centering
        \includegraphics[width=\linewidth, height=4.45cm]{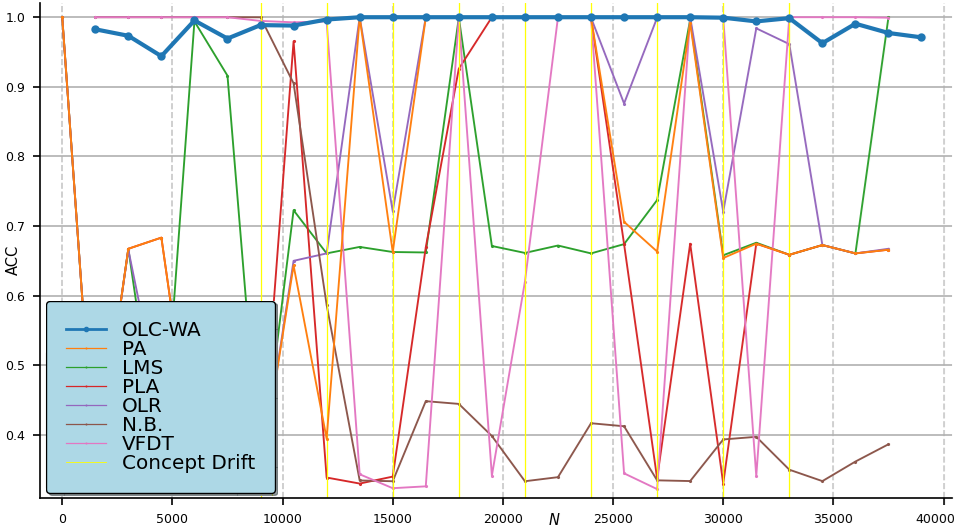}
        \vspace{-.45cm}
        \caption{\scriptsize Models' Performance on DS22}
        \label{fig:sub4}
    \end{subfigure}    
    \vspace{-.2cm}
\caption{\footnotesize Models' performance Incremental Concept Drift Datasets \\
\centering}

% Add Reproducibility Parameters Below
\hspace*{40pt}
    \begin{minipage}{0.9\textwidth}
        \tiny
        \textbf{*Reproducibility Parameters:}
        
        \textbf{LMS} ($\lambda$):  
        $\bullet$ [DS19, DS21] \(\rightarrow\) 0.1  
        $\bullet$ [DS20, DS22] \(\rightarrow\) 0.01
        
        \textbf{OLR} ($\lambda$):  
        $\bullet$ [DS19, DS20, DS21, DS22 ] \(\rightarrow\) 0.1
        
        \textbf{PA} ($C$):  
        $\bullet$ [DS21]  \(\rightarrow\) 1  
        $\bullet$ [DS19, DS20, DS22] \(\rightarrow\)  0.1
        
        \textbf{OLC-WA}:  
        $\bullet$ $\alpha = 0.5$ $\bullet$ KPI = ACC $\bullet$ $\rho$ $\in [0.06, 0.006]$
        $\bullet$ $\zeta$ = 0.005
    \end{minipage}
    
\label{fig:exp_incremental}
\end{figure*}

\textit{Performance Evaluation:} 
The models were tested on the DS19 to DS22 incremental drift datasets (Figure~\ref{fig:exp_incremental}). \modelname consistently achieved the most effective adaptation to incremental drift. In contrast, VFDT and Naive Bayes showed limited ability to cope with changes. PA performed competitively but was highly sensitive to the regularization parameter: tuning it for faster drift response degraded its stability in more stationary settings, as seen in DS19. Overall, all models, including PA, exhibited sub-optimal performance in handling incremental drift.

\begin{figure*}[!htbp]
    \centering
    \begin{subfigure}{0.48\textwidth}
        \centering
        \includegraphics[width=\linewidth, height=4.5cm]{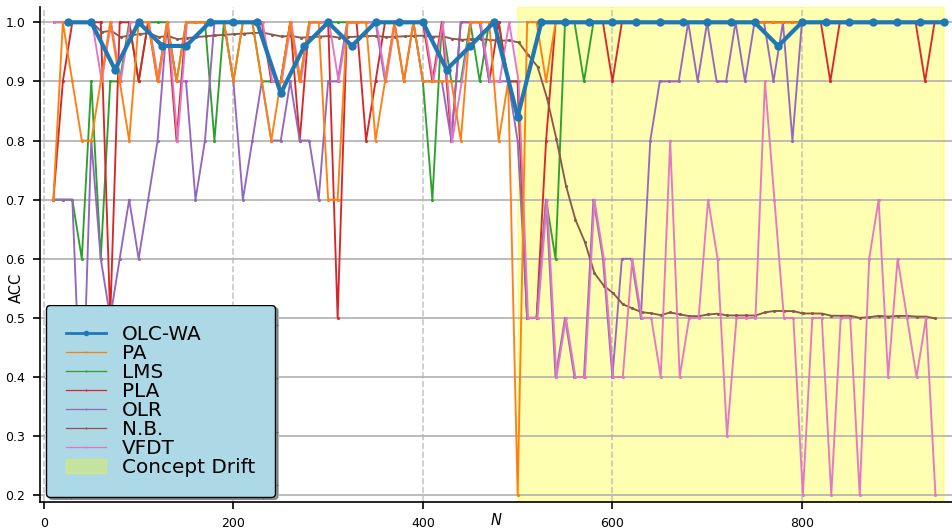}
        \vspace{-.45cm}
        \caption{\scriptsize Models' Performance on DS15}
        \label{fig:sub1}
    \end{subfigure}
    \hfill
    \begin{subfigure}{0.48\textwidth}
        \centering
        \includegraphics[width=\linewidth, height=4.5cm]{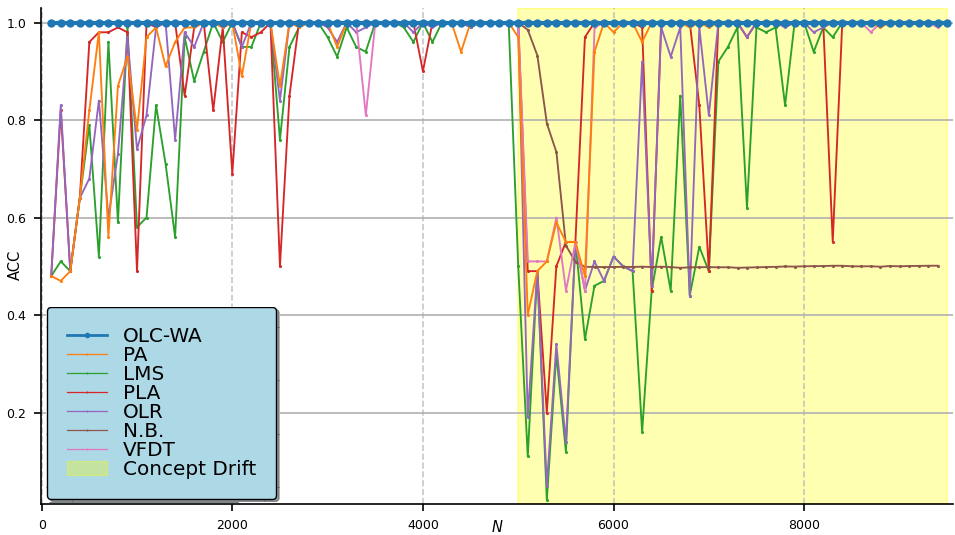}
        \vspace{-.45cm}
        \caption{\scriptsize Models' Performance on DS16}
        \label{fig:sub2}
    \end{subfigure}    
    \begin{subfigure}{0.48\textwidth}
        \centering
        \includegraphics[width=\linewidth, height=4.5cm]{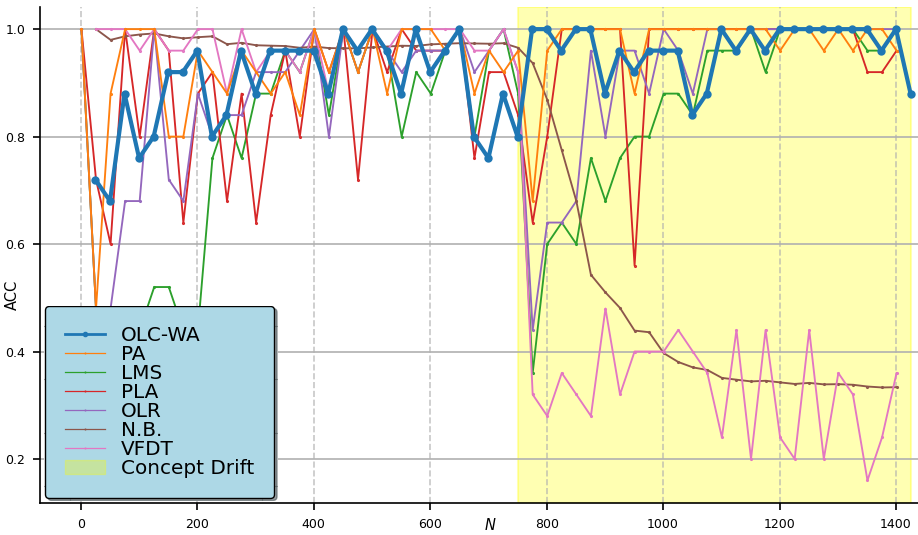}
        \vspace{-.45cm}
        \caption{\scriptsize Models' Performance on DS17}
        \label{fig:sub3}
    \end{subfigure}
    \hfill
    \begin{subfigure}{0.48\textwidth}
        \centering
        \includegraphics[width=\linewidth, height=4.5cm]{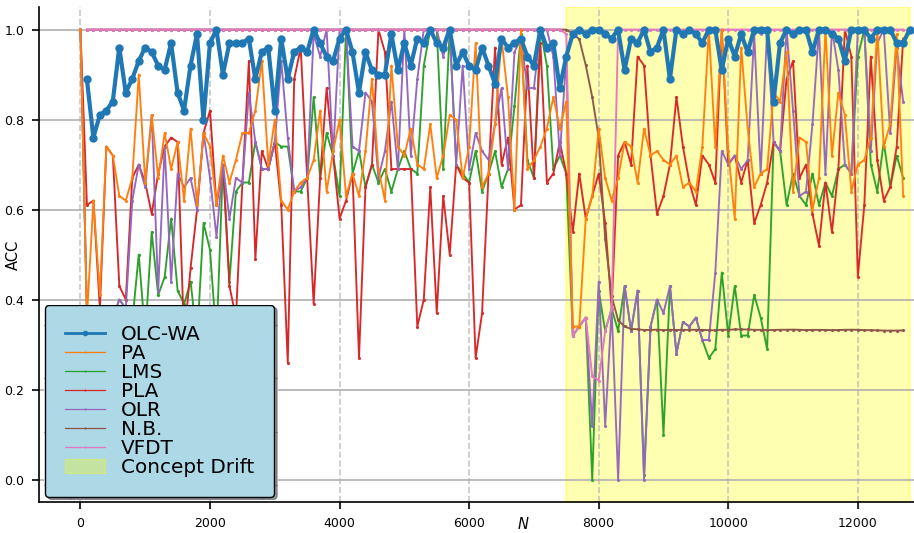}
        \vspace{-.45cm}
        \caption{\scriptsize Models' Performance on DS18}
        \label{fig:sub4}
    \end{subfigure}    
    \vspace{-.2cm}
 \caption{\footnotesize Models' performance Abrupt Concept Drift Datasets \\
\centering}
    \label{fig:exp_abrupt}
% Add Reproducibility Parameters Below
    \hspace*{50pt} % adjust value as needed
    \begin{minipage}{0.9\textwidth}
        \tiny        
        \textbf{*Reproducibility Parameters:}
        
        \textbf{LMS} ($\lambda$):  
        $\bullet$ [DS15] \(\rightarrow\) 0.1  
        $\bullet$ [DS16, DS17] \(\rightarrow\) 0.01  
        $\bullet$ [DS18] \(\rightarrow\) 0.001  
        
        \textbf{OLR} ($\lambda$):  
        $\bullet$ [DS15, DS16, DS17] \(\rightarrow\) 0.1  
        $\bullet$ [DS18] \(\rightarrow\) 0.01
        
        \textbf{PA} ($C$):  
        $\bullet$ [DS15, DS16, DS17]  \(\rightarrow\) 1  
        $\bullet$ [DS18] \(\rightarrow\)  0.1
        
        \textbf{OLC-WA}:  
        $\bullet$ $\alpha$ = 0.5  $\bullet$ KPI = ACC $\bullet$ $\rho$ $\in [0.06, 0.006]$
        $\bullet$ $\zeta$ = 0.005
    \end{minipage}

\label{fig:exp_abrupt}
\end{figure*}

\begin{figure*}[!htbp]
    \centering
    \begin{subfigure}{0.48\textwidth}
        \centering
        \includegraphics[width=\linewidth, height=4.45cm]{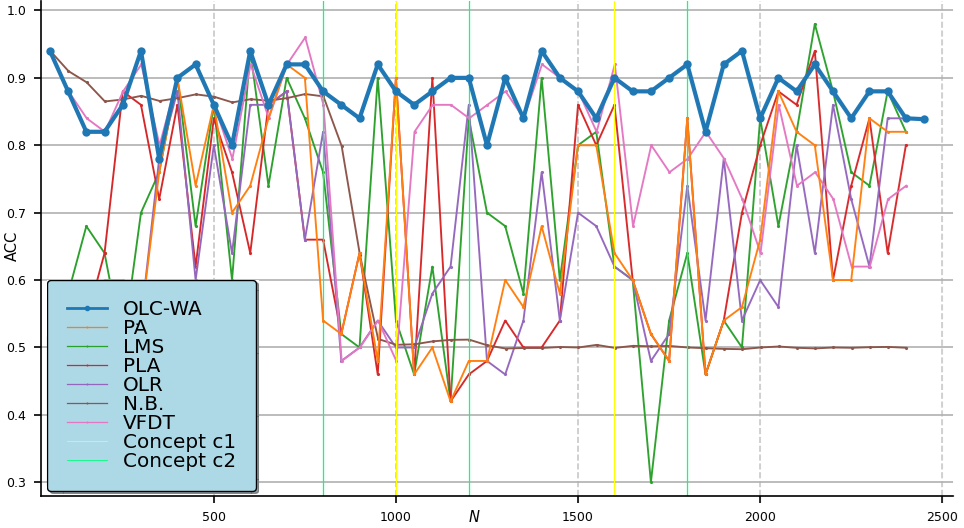}
        \vspace{-.45cm}
        \caption{\scriptsize Models' Performance on DS23}
        \label{fig:sub1}
    \end{subfigure}
    \hfill
    \begin{subfigure}{0.48\textwidth}
        \centering
        \includegraphics[width=\linewidth, height=4.45cm]{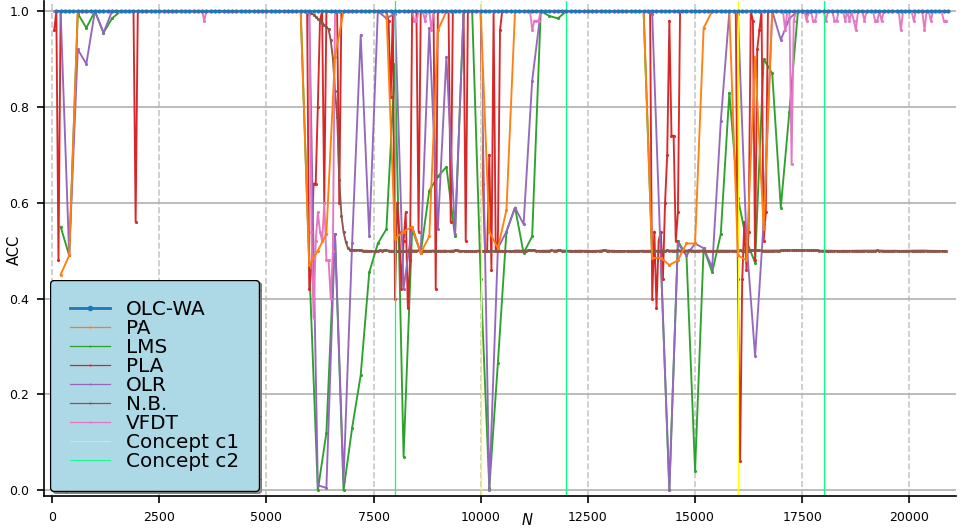}
        \vspace{-.45cm}
        \caption{\scriptsize Models' Performance on DS24}
        \label{fig:sub2}
    \end{subfigure}    
    \begin{subfigure}{0.48\textwidth}
        \centering
        \includegraphics[width=\linewidth, height=4.45cm]{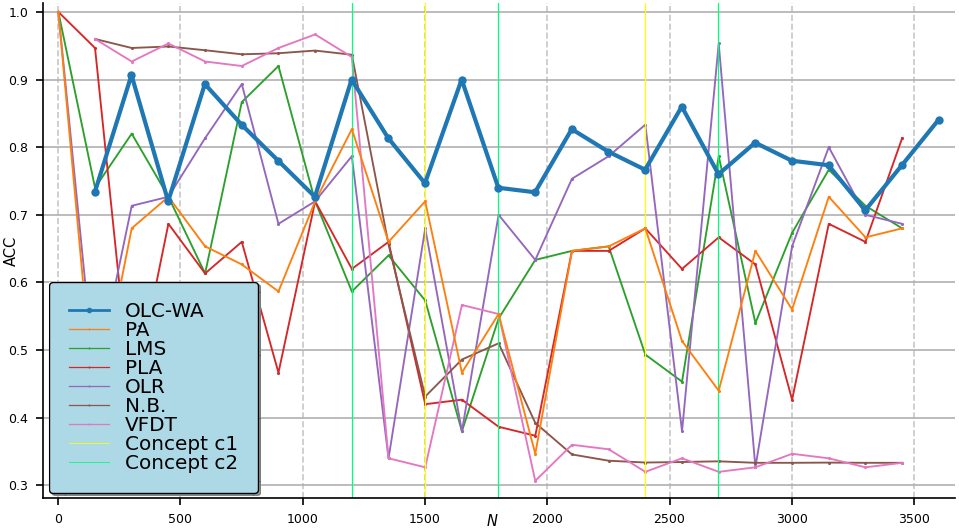}
        \vspace{-.45cm}
        \caption{\scriptsize Models' Performance on DS25}
        \label{fig:sub3}
    \end{subfigure}
    \hfill
    \begin{subfigure}{0.48\textwidth}
        \centering
        \includegraphics[width=\linewidth, height=4.45cm]{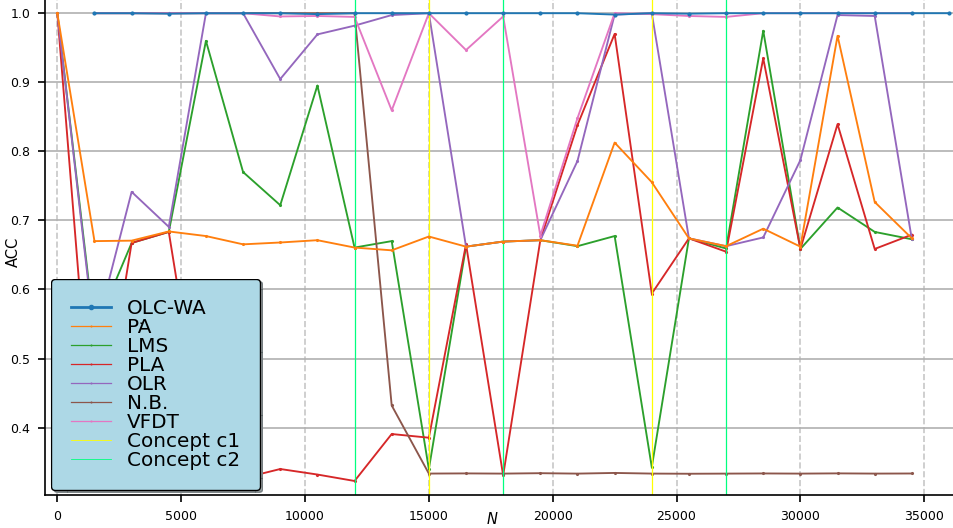}
        \vspace{-.45cm}
        \caption{\scriptsize Models' Performance on DS26}
        \label{fig:sub4}
    \end{subfigure}    
    \vspace{-.2cm}
\caption{\footnotesize Models' performance Gradual Concept Drift Datasets \\
\centering}
% Add Reproducibility Parameters Below
\hspace*{47pt}
    \begin{minipage}{0.9\textwidth}
        \tiny
        \textbf{*Reproducibility Parameters:}
        
        \textbf{LMS} ($\lambda$):  
        $\bullet$ [DS23, DS25] \(\rightarrow\) 0.1  
        $\bullet$ [DS24, DS26] \(\rightarrow\) 0.01
        
        \textbf{OLR} ($\lambda$):  
        $\bullet$ [DS23, DS24, DS25, DS26] \(\rightarrow\) 0.1
        
        \textbf{PA} ($C$):  
        $\bullet$ [DS23, DS25]  \(\rightarrow\) 1  
        $\bullet$ [DS24, DS26] \(\rightarrow\)  0.1
        
        \textbf{OLC-WA}:  
        $\bullet$ $\alpha = 0.5$ $\bullet$ KPI = ACC $\bullet$  $\rho$ $\in [0.06, 0.006]$
        $\bullet$ $\zeta$ = 0.005
    \end{minipage}
\label{fig:exp_gradual}
\end{figure*}

\subsubsection{Gradual Drift}
\textit{Performance Evaluation}: 
The models were evaluated on the DS23 through DS26 gradual drift datasets (Figure~\ref{fig:exp_gradual}). Overall, model performance was below expectations. However, \modelname distinguished itself by delivering superior results, reinforcing its robustness in handling gradual concept drift. Its effective adaptation to evolving data distributions underscores its strength in maintaining stable and reliable performance over time.

\subsubsection{Empirical Runtime}
While the theoretical complexity of the models has been previously analyzed, this subsection presents an empirical evaluation of wall-clock execution time for each algorithm across multiple datasets. Practical runtime behavior often reflects additional factors beyond theoretical bounds, including implementation overhead and system-level operations such as I/O latency. Therefore, this empirical assessment is essential for quantifying the computational efficiency of \modelname\ relative to competing online learning algorithms. The aggregated results are summarized in Table~\ref{tab:runtime_analysis}.

\begin{table}[H]
\centering
\setlength{\tabcolsep}{4pt}
\renewcommand{\arraystretch}{1.1}
\caption{\centering Runtime Analysis.\\
{\scriptsize Average wall-clock time (s) from 5-fold cross-validation with seed averaging.\\ Reproducibility parameters follow Table~\ref{tab:algorithm-performance}.}}
\vspace{-.2cm}
\label{tab:runtime_analysis}
\begin{adjustbox}{width=\textwidth}
\begin{tabular}{lcccccccc}
\toprule
\textbf{Dataset} & \textbf{Batch} & \textbf{PLA} & \textbf{LMS} & \textbf{OLR} & 
\textbf{ONB} & \textbf{PA} & \textbf{VFDT} & \textbf{\modelname}\\
\midrule
\textbf{DS1}   & 0.00169 & 0.00248 & 0.00261 & 0.00366 & 0.80134 & 0.58993 & 0.03856 & 0.15124\\
\textbf{DS2}   & 0.02220 & 0.09717 & 0.09950 & 0.14801 & 200.31903 & 25.65530 & 3.96184 & 0.65972\\
\textbf{DS3}   & 0.04758 & 0.13667 & 0.14141 & 0.18172 & 200.52084 & 54.21052 & 27.31387 & 0.23869\\
\textbf{RSIND} & 0.00647 & 0.01001 & 0.00943 & 0.01346 & 2.37244 & 1.41017 & 0.22115 & 0.18503\\
\textbf{CCDD}  & 0.01757 & 0.00980 & 0.01223 & 0.01369 & 4.46785 & 2.08307 & 0.64412 & 0.11972\\
\textbf{ESPD}  & 0.02508 & 0.01808 & 0.01756 & 0.02779 & 15.25138 & 6.05427 & 2.36936 & 0.28136\\
\textbf{DS7}   & 0.00929 & 0.01557 & 0.02077 & 0.05384 & 2.60236 & 3.07834 & 0.24513 & 0.24673\\
\textbf{DS8}   & 0.05591 & 0.10282 & 0.19162 & 0.46125 & 175.19794 & 38.88421 & 7.97142 & 1.56227\\
\textbf{DS9}   & 0.12320 & 0.17667 & 0.30887 & 0.62805 & 146.77729 & 55.97707 & 78.12476 & 2.82672\\
\textbf{CSCR}  & 0.16000 & 0.21000 & 0.32000 & 0.70000 & 122.00000 & 38.00000 & 62.00000 & 2.95000\\
\textbf{HARD}  & 1.17936 & 0.25160 & 0.43691 & 1.22130 & 167.19299 & 123.67010 & 158.08317 & 5.78082\\
\textbf{HDWD}  & 0.17414 & 0.04587 & 0.10188 & 0.20177 & 8.98475 & 9.12818 & 6.39191 & 0.28674\\
\bottomrule
\end{tabular}
\end{adjustbox}
\end{table}

The empirical results in Table~\ref{tab:runtime_analysis} confirm the expected trend of execution time increasing proportionally with dataset size and dimensionality. As a result, smaller datasets (e.g., DS1, RSIND) complete within milliseconds, while high-dimensional datasets (e.g., DS8, DS9) require several seconds.

$\modelname$ achieves a superior $\modelname$ achieves a superior performance-to-efficiency trade-off within the online learning category, demonstrating high efficiency relative to its adaptive complexity. While simpler linear models are faster, their use is limited by their poor and unstable performance in concept drift scenarios. Conversely, the more competitive adaptive methods VFDT, PA, and ONB incur significant computational overhead. $\modelname$ is consistently substantially faster than these complex models, with its runtime being up to two orders of magnitude lower than that of ONB and PA on larger datasets. This combination of high adaptability and practical efficiency confirms $\modelname$'s strong competitive position and suitability for real-time applications.

\section{Discussion}

\modelname introduces an adaptive, hyperparameter-free framework for online classification that effectively balances stability and adaptability in dynamic streaming environments. This section presents key discussion points along with the model’s advantages, limitations, and potential practical applications.

\subsection{Realization of the Learning Objective Function} 
The learning objective function, introduced earlier, defines how \modelname balances stability and adaptability during online updates. 
It is expressed as:
\begin{multline*}
\mathbf{w}_{t+1} = \underset{\mathbf{w}}{\arg\min} \Bigg( 
    \frac{1}{2} \| \mathbf{w} - \mathbf{w}_t \|^2 + \alpha \mathcal{L}(y_{\text{inc}}, f(\mathbf{w}; \mathbf{X}_{\text{inc}})) \\[-0.5cm]
    + (1 - \alpha) \mathcal{L}(y_{\text{base}}, f(\mathbf{w}; \mathbf{X}_{\text{base}})) \Bigg)
\end{multline*}

Minimization of this objective is performed online through a closed-form proximal update, which results in an Exponentially Weighted Moving Average (EWMA) of the norm vectors:
$
V_{\text{avg}} = (1 - \alpha)\,V_{\text{base}} + \alpha\,V_{\text{inc}}.
$
The vectors \(V_{\text{base}}\) and \(V_{\text{inc}}\) are obtained by minimizing their respective loss functions, 
\(\mathcal{L}_{\text{base}} = \mathcal{L}(y_{\text{base}}, f(\mathbf{w}; \mathbf{X}_{\text{base}}))\) and 
\(\mathcal{L}_{\text{inc}} = \mathcal{L}(y_{\text{inc}}, f(\mathbf{w}; \mathbf{X}_{\text{inc}}))\),
over the base and incremental mini-batches. 
Hence, the loss weights \((\alpha, 1-\alpha)\) directly determine the weighting of their corresponding norm vectors in the parameter update.

When a \textbf{severe (abrupt) drift} is detected, \modelname assigns a high value to \(\alpha\) (typically close to 1). 
In this case, the second term, \(\alpha\,\mathcal{L}(y_{\text{inc}}, f(\mathbf{w}; \mathbf{X}_{\text{inc}}))\), dominates the objective function. 
This shifts the optimization focus toward minimizing the loss on the most recent mini-batch, causing the updated parameter vector \(\mathbf{w}_{t+1}\) to move rapidly in the direction of the incremental norm vector \(V_{\text{inc}}\). 
As a result, the model adapts quickly to the new concept, effectively re-aligning the decision boundary with the latest data distribution.

Conversely, when the detected drift is \textbf{minor (incremental)} or the stream remains stable, \(\alpha\) is set to a moderate or smaller value (e.g., \(\alpha \leq 0.5\)). 
Here, the first term, \(\tfrac{1}{2}\|\mathbf{w}-\mathbf{w}_t\|^2\), and the third term, \((1-\alpha)\mathcal{L}(y_{\text{base}}, f(\mathbf{w}; \mathbf{X}_{\text{base}}))\), dominate the objective. 
This configuration penalizes large parameter shifts and places greater emphasis on maintaining consistency with historical data. 
Consequently, the update remains close to \(V_{\text{base}}\), enhancing stability and preventing overreaction to minor fluctuations or noise.

Through this adaptive adjustment of \(\alpha\), \modelname consistently minimizes the same objective across all data streams, dynamically emphasizing either adaptability or stability based on the detected drift severity.

\subsection{Geometric Robustness and Intersection Handling}
\label{sec:intersection_handling}

The \modelname\ framework relies fundamentally on the existence of an intersection point between the base and incremental hyperplanes to define the geometric transition between the two decision boundaries. Under continuous-valued features and stochastic updates (e.g., mini-batch sampling), such an intersection exists with probability one. When an intersection is not found within numerical tolerance, the configuration reduces to one of two degenerate cases: the \textit{Coincide} case or the \textit{Parallel} case.

% \begin{enumerate}[label=(\roman*), leftmargin=2em]
\begin{enumerate}[labelsep=0.8em]
\item \textbf{Coincide:} The base and incremental decision boundaries are identical, i.e., $\mathbf{V}_{\text{base}} = \mathbf{V}_{\text{inc}}$ and $b_{\text{base}} = b_{\text{inc}}$. In this case, the decision surfaces overlap entirely, indicating that the incremental update introduces no new information and thus no adjustment is necessary.

\item \textbf{Parallel:} The two hyperplanes share the same orientation ($\mathbf{V}_{\text{inc}} = c \mathbf{V}_{\text{base}}$ for some nonzero scalar $c$) but differ in offset. Although exact parallelism occurs only when $\lvert \cos \theta \rvert = 1$, where $\theta$ is the angle between the normal vectors, nearly parallel configurations ($\lvert \cos \theta \rvert \approx 1$) result in ill-conditioned intersection computations. To ensure robustness, \modelname\ replaces the intersection point with a \textit{weighted midpoint} between the two planes. The weighting factor $\alpha$ determines the location: $\alpha = 0.5$ yields the geometric midpoint, $\alpha > 0.5$ moves the update toward the incremental boundary, and $\alpha < 0.5$ favors the base boundary. This safeguard guarantees a well-defined update even in edge cases, though exact parallelism remains theoretically improbable under continuous-valued, stochastic updates~\cite{tao2011introduction, bottou2018optimization, de2008computational}.
\end{enumerate}

\subsection{Extending \modelname to Nonlinear Domains}
\label{sec:nonlinear_discussion}

Extending the \modelname\ framework to nonlinear classification represents a natural and essential direction for future research. Nonlinearity can be introduced by mapping the input data $\mathbf{x}$ into a higher-dimensional feature space $\mathbf{z} = \phi(\mathbf{x})$ through a kernel function, such as the radial basis function (RBF) or polynomial kernel. While this transformation enables the model to capture complex, curved decision boundaries, it simultaneously invalidates the geometric assumptions underlying the linear \modelname\ formulation: the decision surface no longer has a single, consistent norm vector, and the computation of a unique global intersection point becomes ill-posed and numerically unstable.

To address this geometric limitation, the nonlinear extension of \modelname\ should generalize the core adaptive mechanism by replacing parameter-space blending with an exponentially weighted combination applied directly in the \textit{function space}. This modification eliminates the dependence on norm vectors and explicit intersection computation, both of which are undefined for nonlinear surfaces. The proposed approach interpolates the predicted outputs of the base and incremental models, effectively applying the EWMA directly to the decision function:
$
f_{\text{avg}}(\mathbf{x}) = (1 - \alpha) f_{\text{base}}(\mathbf{x}) + \alpha f_{\text{inc}}(\mathbf{x}).
$ 

This functional averaging strategy provides a theoretically consistent nonlinear analog of the \modelname\ update, preserving the framework’s stability–adaptability balance governed by $\alpha$, while ensuring a smooth and globally consistent transition between model states across the input domain. Importantly, the featured drift detection and adaptation mechanism remains unchanged, as it is KPI-dependent and model-agnostic. Future work will focus on the empirical validation of this formulation and its evaluation across diverse nonlinear settings.

\subsection{Configuration Parameters Settings}
\begin{figure*}[!htbp]
    \centering
    \begin{subfigure}{0.48\textwidth}
        \centering
        \includegraphics[width=\linewidth, height=4cm]{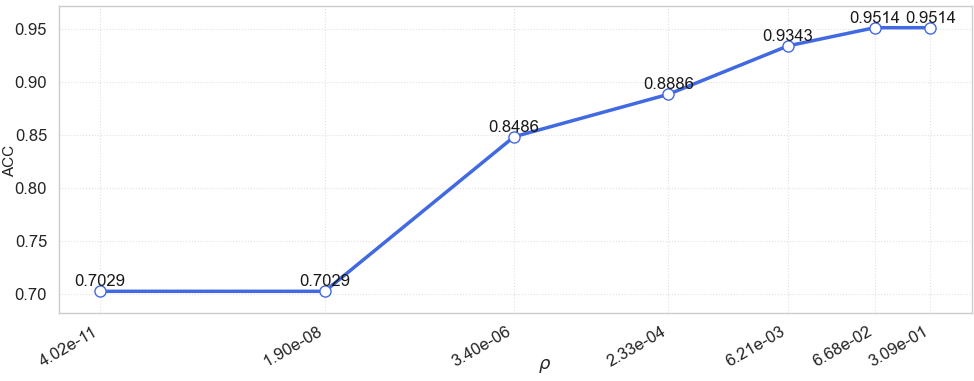}
        \vspace{-.45cm}
        \caption{\scriptsize \modelname sensitivity to $\rho$}
        \label{fig:sub1}
    \end{subfigure}
    \hfill
    \begin{subfigure}{0.48\textwidth}
        \centering
        \includegraphics[width=\linewidth, height=4cm]{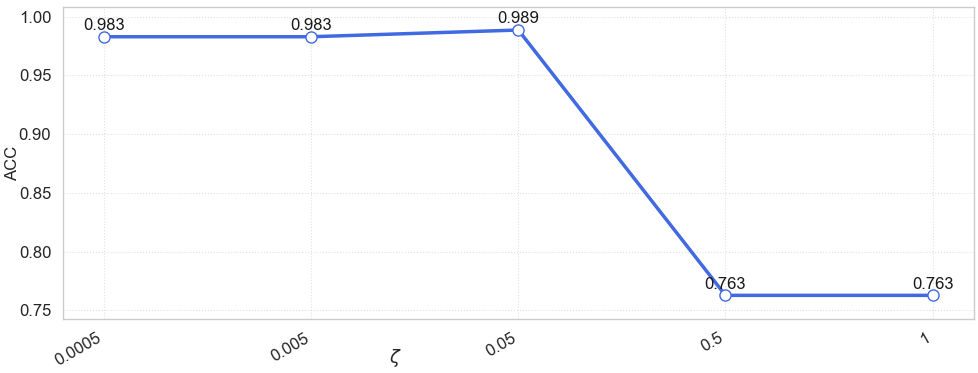}
        \vspace{-.45cm}
        \caption{\scriptsize \modelname sensitivity to $\zeta$}
        \label{fig:sub2}
    \end{subfigure}        
    \vspace{-.2cm}
\caption{\footnotesize \modelname Sensitivity to Configuration Parameters \centering}
\label{fig:sensitivity_parameters}
\end{figure*}

As discussed earlier, \modelname\ is hyperparameter-free, relying only on user-defined, 
non-trainable configuration parameters that remain fixed during deployment. 
This subsection outlines these parameters and their recommended settings.
\begin{enumerate}
    \item \textit{Key Performance Indicator (KPI):} 
    The chosen KPI should be aligned with the specific learning task; for instance, 
    in classification problems, commonly used KPIs include \textit{Accuracy}, \textit{F1-score}, 
    or \textit{Cross-Entropy Loss}. 
    The KPI is employed to evaluate model stability and guide drift detection, serving as the 
    statistical signal from which the mean ($\mu_{\text{KPI}}$) and variance ($\sigma_{\text{KPI}}$) 
    are computed within the Gaussian-based detection module. 
    In our experiments, \textit{Accuracy} was primarily used as the KPI, 
    as it provides a direct and interpretable measure of predictive consistency for classification tasks. Nevertheless, the framework is agnostic to the specific KPI or set of KPIs employed, 
    and can seamlessly adapt to different metrics based on the application domain.

    \item \textit{KPI-Window Size (KWS):} 
    Defines the number of recent KPI observations used to compute $\mu_{\text{KPI}}$ and $\sigma_{\text{KPI}}$ for drift monitoring. It is computed as $\text{LB} \leq \text{KWS} = \left(\frac{N}{K}\right) \times \gamma \leq \text{UB}$, which keeps the window size bounded and prevents uncontrolled growth over long streams. Larger windows provide smoother estimates but slower response, while smaller ones increase sensitivity. In our experiments, $\text{KWS} = 31$ with $\gamma = 0.05$ offered a balanced trade-off.

    \item \textit{Stability Band Parameter ($\zeta$):} This parameter defines the half-width of the inner stability band 
    $[\mu_{\text{KPI}} - \zeta,\ \mu_{\text{KPI}} + \zeta]$. 
    Its primary function is to absorb minor, noise-level fluctuations in the monitored KPI, 
    thereby preventing the system from generating overreactive updates during stable periods. 
    It explicitly distinguishes noise-level variations from meaningful drift, enhancing the model’s robustness and stability. 
    The specific value of $\zeta$ should be selected according to the chosen KPI and its numerical scale. 
    For example, when \textit{Accuracy} is used (scale $\in [0, 1]$), $\zeta$ was fixed to $0.005$, corresponding to a $\pm0.5\%$ tolerance. This value was determined from the \modelname\ sensitivity analysis shown in Figure~\ref{fig:sensitivity_parameters}(b), which indicates optimal $\zeta$ values around $0.005$.

    \item \textit{False-Alarm Probability ($\rho$):} 
    This parameter controls the confidence level of the Gaussian-based drift detection module 
    and defines the probability of incorrectly classifying normal KPI fluctuations as drift events. 
    Smaller values of $\rho$ make the detector more conservative, reacting only to statistically 
    significant deviations, whereas larger values increase sensitivity to minor performance changes. 
    Mathematically, $\rho$ determines the cutoff threshold through the inverse Gaussian CDF 
    $z = \Phi^{-1}(1 - \rho)$, where $\Phi^{-1}(\cdot)$ is the standard normal quantile function, and 
    the resulting threshold $\tau = z\sigma$ scales adaptively with the KPI variance.

    In our experiments, the $\rho$ values were derived from the \modelname\ sensitivity analysis to $\rho$, as illustrated in Figure~\ref{fig:sensitivity_parameters}(a). The analysis identified the optimal operational range of $\rho$ between $0.006$ and $0.06$, corresponding to approximately $2.5\sigma$–$1.5\sigma$ confidence thresholds. This range provides a balanced trade-off between sensitivity and stability. For instance, selecting a higher false-alarm probability ($\rho = 0.06$) makes the detector more sensitive, allowing it to classify subtle performance changes as potential drift, whereas choosing a lower probability ($\rho = 0.006$) yields a more conservative behavior, treating minor fluctuations as stable.

    The parameter $\rho$ is independent of the specific KPI selected, as it operates on the 
    statistical distribution of the KPI rather than its absolute value. 
    This ensures that drift detection sensitivity is governed by standardized deviations 
    from the KPI’s mean, not by the metric’s numerical scale.
\end{enumerate}

%HEREMOHA

\subsection{Advantages}
The primary advantage of \modelname\ lies in its \textbf{proactive drift awareness} and \textbf{tuning-free design}, supported by its integrated drift detection and adaptation mechanism. In contrast to traditional online learning models that rely on frequent manual hyperparameter optimization and external reactive drift detectors, which identify drift only after it occurs, \modelname\ automatically adjusts its smoothing factor $\alpha$ in real time based on observed performance indicators. This enables continuous and autonomous operation under dynamic and evolving data conditions.

\subsection{Limitations}
The adaptation behavior of $\text{\modelname}$ is inherently dependent on the quality of the selected Key Performance Indicator ($\text{KPI}$). If the chosen KPI is \textbf{noisy, delayed, or poorly aligned} with the true learning objective, the drift detection and subsequent adjustment of $\alpha$ may respond with latency or reduced sensitivity. This limitation reflects a broader trade-off inherent in all performance-driven adaptation systems: the reliability of adaptation directly mirrors the reliability of the feedback signal. However, $\text{\modelname}$ incorporates specific \textbf{safeguards} to maintain stability and robustness despite this dependency:
\begin{enumerate}
    \item \textit{Statistical Vetting via Constant False-Alarm Rate (CFAR):} Rather than relying on raw KPI scores or arbitrary fixed thresholds, $\text{\modelname}$ employs a statistically rigorous approach that sets the drift limit ($\tau$) based on a user-specified \textbf{false-alarm probability} ($\rho$), derived as $z = \Phi^{-1}(1 - \rho)$. As detailed in \textit{Mitigating False Positives}, this forms a CFAR threshold ensuring that only deviations exceeding a statistically significant level trigger a drift signal. This effectively filters out noise, random fluctuations, and non-significant variations, even in high-variance KPIs.
    
    \item \textit{Robustness via Multi-KPI Utilization and Voting:} The design of the $\text{KPI-Win}$ module is engineered to allow more than one KPI to be utilized simultaneously. To ensure robustness and avoid the noisiness or bias from a single KPI, a combination of multiple KPIs can be employed. In this scenario, the classification of drift severity (e.g., \textit{no drift}, \textit{minor}, or \textit{severe drift}) is determined by a \textbf{majority voting mechanism} across the outputs of the individual KPI-Win modules. This ensemble approach effectively averages out noise and spurious fluctuations present in any single metric, ensuring a drift signal is only acted upon when confirmed by multiple, independent performance indicators.
\end{enumerate}

In addition, the (EWMA) mechanism used in $\text{\modelname}$ may appear, in principle, to introduce a trade-off between stability and adaptability. A standard EWMA employs a \textbf{fixed decay rate}, which is inherently constrained by this trade-off: a slow decay favors stability but limits adaptability, while a fast decay improves adaptability at the cost of stability. However, this potential limitation is mitigated in $\text{\modelname}$ through the use of an \textbf{adaptive smoothing factor} $\alpha_t$, which dynamically adjusts according to the detected drift severity and the statistical characteristics of the observed KPIs. When the data stream remains stable, $\alpha_t$ decreases to preserve historical knowledge and ensure stable learning updates. Conversely, during abrupt or significant drifts, $\alpha_t$ increases, prioritizing faster adaptation to changing patterns. This adaptive control of the EWMA decay effectively rebalances stability and adaptability in real time, ensuring that $\text{\modelname}$ remains responsive yet robust under varying drift intensities.

\subsection{Extensibility to Complex ML Tasks}

The current version of \modelname focuses on online classification in non-stationary environments, which is a critical and complex machine learning task in its own right. Nevertheless, the framework is conceptually general and can be extended to more complex tasks. For instance, it can be extended to tasks such as multi-label classification, multi-output regression, and object detection. 

For multi-label classification, each label can be modeled as an independent binary task, each with its own base and incremental classifiers and adaptive weighting factor ($\alpha^{(k)}$). This allows independent adaptation to drift based on label-specific performance. Similarly, in multi-output regression, the weighted averaging and intersection mechanisms can be applied independently to each output component, ensuring stable, drift-aware updates for every target variable. For object detection, the adaptive update rule can be applied to the classification and localization heads independently, allowing the network to adapt to shifts in object appearance (classification drift) and bounding box scales (regression drift) online. Furthermore, in deep or hierarchical architectures, \modelname's adaptive update rule (controlled by $\alpha_t$) can be incorporated at the layer or module level to enable fine-grained online adaptation without fully retraining the network, making it suitable for continual learning. While these extensions require some architectural modifications to manage complex or high-dimensional outputs, the core principles of \modelname remain applicable. The adaptive weighting mechanism ($\alpha_t$) and the weighted combination of representations provide a flexible foundation for adaptation across various complex tasks.

\subsection{Application Features and Prospects}

\modelname is well-suited for practical applications where models must learn and adapt continuously with minimal human intervention. The framework has been empirically validated across multiple real-world datasets spanning diverse domains, including customer support text streams, sensor-based monitoring, and financial transaction data, demonstrating consistent adaptability under various drift patterns. Its hyperparameter-free nature makes it especially suitable for Machine Learning as a Service (MLaaS) deployments, where user-specific parameter tuning is typically infeasible and models must autonomously adjust to drifty or evolving data streams. Moreover, the framework’s explainable drift-handling mechanism enhances transparency in decision-critical environments, supporting seamless integration into production-grade online learning systems.

\section{Conclusion and Future Work}

\modelname as an adaptive, tunable-free model with a built-in optimization technique, offers solutions to several challenges. 
First, \modelname, as a drift-aware model, features a proactive in-memory built-in drift detection and adaptation mechanism, which employs a variable threshold technique and uses key performance indicators (KPIs) without assuming any specific data distribution, making it particularly suited for dynamic and unpredictable environments. 
Second, the dynamic nature of the data makes hyperparameters in online models impractical, as continuous manual adjustments become unmanageable and potentially infeasible. \modelname, as a hyperparameter-free model, overcomes this limitation and offers flexibility in various scenarios. 
Third, \modelname mitigates overfitting and catastrophic forgetting by incorporating an EWMA decay mechanism, and by dynamically adjusting adaptive hyperparameters in real time, enabling learning from new data while preserving past knowledge. 
Fourth, \modelname explicitly manages false positives through a user-specified false-alarm probability $\rho$, which calibrates the detection threshold to guarantee a constant false-positive rate. This design choice gives practitioners direct control over the trade-off between sensitivity to subtle drifts and robustness against spurious fluctuations. 
Fifth, \modelname can seamlessly convert batch classifiers into online learning models, providing a practical pathway for adapting existing methods to dynamic data stream environments. 
Empirical evaluation against a suite of state-of-the-art online classification models demonstrates that \modelname excels in both convergence and its capability to effectively manage various types of concept drift. 
Future work includes extending \modelname to broader ML tasks and deploying it as an online web service for drift-aware real-time classification across diverse domains.

%###########################################################################################

%% The Appendices part is started with the command \appendix;
%% appendix sections are then done as normal sections
% \appendix
% \section{Example Appendix Section}
% \label{app1}

% Appendix text.

% %% For citations use: 
% %%       \cite{<label>} ==> [1]

% %%
% Example citation, See \cite{lamport94}.

%% If you have bib database file and want bibtex to generate the
%% bibitems, please use
%%
%%  \bibliographystyle{elsarticle-num} 
%%  \bibliography{<your bibdatabase>}

%% else use the following coding to input the bibitems directly in the
%% TeX file.

%% Refer following link for more details about bibliography and citations.
%% https://en.wikibooks.org/wiki/LaTeX/Bibliography_Management

% \begin{thebibliography}{00}

% %% For numbered reference style
% %% \bibitem{label}
% %% Text of bibliographic item

% \bibitem{lamport94}
%   Leslie Lamport,
%   \textit{\LaTeX: a document preparation system},
%   Addison Wesley, Massachusetts,
%   2nd edition,
%   1994.

% \end{thebibliography}

\bibliographystyle{elsarticle-num}  % numeric style (common for ESWA)
\bibliography{references}           % references.bib file name (no .bib extension)

\end{document}